# Developing a Free and Open-source Automated Building Exterior Crack Inspection Software for Construction and Facility Managers

Pi Ko [a], Samuel A. Prieto [a], Borja García de Soto [a]

[a] S.M.A.R.T. Construction Research Group, Division of Engineering, New York University Abu Dhabi (NYUAD), Experimental Research Building, Saadiyat Island, P.O. Box 129188, Abu Dhabi, United Arab Emirates

E-mail: pk2269@nyu.edu, samuel.prieto@nyu.edu, garcia.de.soto@nyu.edu

https://github.com/SMART-NYUAD/ABECIS

**Abstract**

Inspection of cracks is an important process for properly monitoring and maintaining a building. However, manual crack inspection is time-consuming, inconsistent, and dangerous (e.g., in tall buildings). Due to the development of open-source AI technologies, the increase in available Unmanned Aerial Vehicles (UAVs) and the availability of smartphone cameras, it has become possible to automate the building crack inspection process.

This study presents the development of an easy-to-use, free and open-source Automated Building Exterior Crack Inspection Software (ABECIS) for construction and facility managers, using state-of-the-art segmentation algorithms to identify concrete cracks and generate a quantitative and qualitative report. ABECIS was tested using images collected from a UAV and smartphone cameras in real-world conditions and a controlled laboratory environment. From the raw output of the algorithm, the median Intersection over Unions for the test experiments is (1) 0.686 for indoor crack detection experiment in a controlled lab environment using a commercial drone, (2) 0.186 for indoor crack detection at a construction site using a smartphone and (3) 0.958 for outdoor crack detection on university campus using a commercial drone. These IoU results can be improved significantly to over 0.8 when a human operator selectively removes the false positives. In general, ABECIS performs best for outdoor drone images, and combining the algorithm predictions with human verification/intervention offers very accurate crack detection results.

The software is available publicly and can be downloaded for out-of-the-box use at: https://github.com/SMART-NYUAD/ABECIS

## 1. Introduction

Depending on the maintenance plan, the inspection of buildings is a time-consuming, repeating process with intervals between 5 to 10 years [1]. For the most part, inspections are done manually by inspectors. It is a labor-intense task when done for large buildings (or a portfolio of buildings). Moreover, it is dangerous if the inspection on the exterior of the building is required for high-rise buildings because traditionally, it requires an inspector to abseil down over different sides of a building [2]. Also, the manual visual examination is usually done by experienced operators with specialized surveying tools such as magnifiers, crack rulers, etc. Sometimes, the results can be inconsistent and suffer from human subjectivity [3]. Therefore, the manual inspection does not leave consistent computerized data (i.e., digital data), which can be used to compare the results of successive inspections over time later.

In recent years, there have been developments in commercial Unmanned Aerial Vehicles (UAVs) or drones and the widespread availability of smartphones. Moreover, there are also immense developments in open-source Artificial Intelligence (AI) systems. Therefore, it has now become possible to fuse all these technologies to implement a partially, if not fully, autonomous crack detection system which takes digital images from smartphone cameras, UAVs, or robots as an input, then uses AI and image processing techniques to identify the location of the cracks and provides detailed quantitative and qualitative reports. There has been an increase in research utilizing image processing techniques and AI to detect and classify cracks with varying levels of success [4]. Nevertheless, many of these research experiments tend to be very technical or research-oriented and have not been adopted by mainstream construction and facility managers. This research aims to close the gap between theoretical research in crack detection and practical application in the field by developing an Automated Building Exterior Crack Inspection Software (ABECIS). ABECIS has been developed as an easy-to-use, free and open-source graphical crack detection software which runs cross-platform on all operating systems. The software only takes the images of buildings (either taken by smartphone cameras, UAVs, or other robotic platforms) as an input, detects the cracks based on user-adjusted parameters and automatically generates detailed quantitative and qualitative reports. Therefore, the construction and facility managers can simply select a folder with all the images of the building they want to inspect using ABECIS, and the rest of the process will be automated with minimum need for human intervention, allowing them to harness all the advantages of AI without the need to be technical or knowledgeable about it. ABECIS has been made publicly available through a GitHub repository [5]. Installation instructions can be found in [6].

In the first part of this paper, an extensive literature review is done through relevant research papers in recent years, as well as a review of publicly available crack detection algorithms online and some industry research papers. Secondly, the paper addresses the necessary background information on data collection with smartphones and drones, classification of building crack types, deep learning, and instance segmentation algorithms. The third part of this paper discusses the research methodology, lays out the theoretical assumptions, and explains the basic functionality of the software. This is followed by an implementation section, where the process of developing ABECIS, its functionalities and outputs are described. Afterward, the results from testing ABECIS with images taken (a) using a commercial drone in a laboratory environment with a mockup concrete wall, (b) using a smartphone camera in an actual construction site and real urban settings and (c) using a commercial drone outdoors on the university campus are evaluated and discussed. Finally, the conclusion, outlook, and acknowledgment are provided.

## 2. Literature Review

A literature review was done through a wide variety of conferences and journals, including but not limited to ISARC proceedings, Automation in Construction, Structure and Infrastructure Engineering, Energies and Sensors, to identify and summarize the state-of-the-art crack detection techniques and their situation at the time of this study. To focus the paper on the application and usability of these techniques on-site in the real world, this literature review discusses academic research papers (subsection 2.1) and addresses the open-source code for crack detection that are publicly available (subsection 2.2).

### 2.1. Literature Review of Academic research

A comprehensive summary of relevant papers addressing the use of image processing techniques (with and without the use of AI), with varying degrees of automation to detect defects on buildings or structures with various surface materials, published in recent years, were identified using a keyword-based search. This study identified the state-of-the-art research trends in crack detection and building monitoring techniques in construction in general, using the following keywords: "building inspection," "automation," "deep learning," "machine learning," "crack detection," "building monitoring," "segmentation," "image processing." The sources are selected based on clarity, the novelty in the methods presented, and their relevance to this study. Findings from that review are provided below, and a summary table with all papers considered is available in Appendix A (Table 5).

From the literature review and a quick inspection of Table 5 in Appendix A, cracks are some of the most common defects detected when inspecting building exteriors. Most of the approaches are semi-autonomous, meaning that at least the data acquisition during the process is manually commanded, or the methodology requires some manual input during the crack recognition process. The most popular platform is a drone (UAV) for data acquisition (i.e., taking images during the inspection).

Similar studies to the one proposed in this paper have been done. However, the proposed method addressed some of the limitations. For example, within the approaches that use Artificial Intelligence (AI) for data processing, a few of the methods [7] [8] [9] use an R-CNN [10] approach for data processing. Ayele et al. [7]

used images of the Skodsberg concrete bridge in Norway taken with a UAV to detect cracks. The length, width and the areas of the cracks were found computationally. However, the study does not provide any datasets or code samples for replicating it. Yu et al. [11] and Cen et al. [12] presented machine learning-based approaches to detect cracks on the concrete surface of bridges. They used a UAV to collect the images to train the algorithm, but those studies do not account for automation in the data collection process. The approach proposed by Yu et al. [11] only provides the classification and localization of the cracks instead of the actual instance segmentation needed to further analyze the crack. In the work of Cen et al. [12], they evaluated how the size of the filter window and the setting of the threshold can affect the results of the crack detection process. However, they only provide a classification of images without a precise localization or instance segmentation of the cracks. Jo et al. [13] presented a system that used multi-layered image processing and deep belief learning to classify road surface cracks. They indicate that their method can be used with a UAV, although they do not test this particular data collection approach, showcasing just RGB and infrared images collected with a smartphone camera (with the help of a FLIR one attachment). The first step in their processing stage was a multi-layered traditional image filtering, applying different filters such as Gabor, Otsu, Retinex and Prewitt filters in order to extract features of segmentation, edges and background. The features were then sent to a Deep Belief Network (DBN) classifier [14]. Their system is only able to classify images, not actually detect and segment the cracks. Within the AI spectrum, CNN-based approaches are the most popular [15] [16] [17] [18] [19] [20]. Silva et al. [15] used YOLOv4 [21] classifier to classify and detect potholes and cracks in the pavement. Both Yeum et al. [16] and Dorafshan et al. [17] used a CNN-based AlexNet approach [22]. Yeum et al. [16] focused on validating Regions of Interest (ROI) around bridge metal joints for further inspection of the welding areas. Dorafshan et al. [17] compared the performance of the CNN approach against traditional image processing filters, proving the superiority of the AI based approaches when it comes to crack detection. Chen et al. [23] proposed a two-step deep learning method for the automated detection of façade cracks. They used a UAV to collect the images and put them through a first classification step where a CNN model was used to classify regions of the image as crack or non-crack. Based on a U-Net neural network model, the second step extracts the pixels corresponding to cracks from the crack-labeled regions. They also compared how their algorithm performs against traditional image processing methods. Despite their method being reliable for crack segmentation, they do not provide any analysis of the detected crack.

Some methods focus on traditional image processing techniques without the involvement of AI techniques. For those focusing on crack detection, the data processing involves multiple algorithms based on edge detection [24] [25] [26] [27] [28] [29] [30] [31]. For the most part, methods based on edge detection cannot provide results as reliable as those with AI data processing. For instance, Liu et al. [26] performed a 3D reconstruction of bridge piers with the help of Structure from Motion (SfM) techniques to later project the located crack in 3D. Although they thoroughly analyzed the detected crack, they did not provide quantitative results on how well the algorithm detected said cracks (i.e., there is no false negatives information). Other approaches do not focus on the precise localization of cracks and rely more on qualitative surface evaluation, such as the absence of paint [32] or variation of defects by measuring the similarity between two pictures [33].

The approaches mentioned above are either manual or semi-autonomous, in the sense that none of them consider the automation of the data collection process. Within the few fully autonomous approaches that have been found, Jo et al. [34] proposed a methodology that would perform fully autonomous inspections on concrete surfaces. The work is based on an autonomous control over the UAV during the data collection, considering different agents such as the wind to control the path of the UAV while both RGB and infrared images are being collected. Both images are later processed by a machine learning agent to identify cracks by combining the detections in both spectrums. Nonetheless, they do not provide any results of the proposed methodology. Kang et al. [35] proposed an ultrasonic beacon system to replace the role of the GPS to provide autonomy to the UAV during the data acquisition. The video data taken with the UAV is later processed with the help of a deep convolutional neural network (CNN) to detect concrete cracks. Their method can successfully locate a small set of cracks present on the concrete floors. However, their approach could only detect and locate cracks without performing actual instance segmentation that could be used to further analyze the crack to provide more data.

Li et al. [36] used an autonomous UAV-based system to detect cracks in mining slopes. They compared some of the most traditional edge detectors (i.e., Prewitt, Sobel, Canny) with a U-Net approach to detect rocks and cracks amongst the slopes. Even though their study highlighted the performance of each method for edge detection, it did not provide an assessment of the cracks detected during the experiment or the outcomes of the project; therefore, it is hard to evaluate the effectiveness of the proposed approach.

**2.2. Industry Research and Available Crack Detection Code**

To consider the development in the industry and the accessibility of these tools in the public domain, the literature review was extended to include code repositories on GitHub and online writings by the industry, such as Canon [37]. The keyword-based search on search engines (Google, DuckDuckGo) and GitHub using the keyword "crack detection software" led to the resources summarized in Table 1.

Table 1. Summary of review on the crack detection code and industry/commercial products available online

| Reference | Type | Code accessible to General Public | For cracks detected on… | Graphical User Interface Available | Dataset available | Pre-trained model available | Generates Report |
|---|---|---|---|---|---|---|---|
| Fan Yang et al. [38] | GitHub | ✓ | Pavement | | ✓ | ✓ | |
| Ruiz et al. [39] | Publication | ✓ | Concrete | | ✓ | | |
| Cha et al. [40] | GitHub | ✓ | Concrete | | ✓ | | |
| Dais et al. [41] | GitHub | ✓ | Masonry | | ✓ | ✓ | |
| Zou et al. [42] | GitHub | ✓ | Pavement | | ✓ | ✓ | |
| Canon [37] | Publication | | Concrete | | | | |
| ABECIS (this study) | Publication and GitHub | ✓ | Concrete | ✓ | ✓ | ✓ | ✓ |

There are existing codes for crack detection available to the public. Except for Zou et al. [42] and Dais et al.[41], that provide their dataset and pre-trained models online so that the public can download their code and run the crack detection algorithms on their own for specific types of cracks (for masonry surface cracks and pavement cracks only), most available code does not offer pre-trained models. Therefore, one must perform the training before the algorithm can be used. Training a machine learning algorithm by oneself has both advantages and limitations. As an advantage, one can fine-tune the parameters, such as the number of epochs, to make the model suitable for the project's specific needs. However, as a limitation, setting up the computing environment and requirements to train the machine learning model may be difficult for unfamiliar individuals. Therefore, everyone cannot utilize this crack detection algorithm unless they can set up the computing environment themselves. It is worth noting that there are several commercially available crack detection algorithms, such as those developed by Dynamic Infrastructure [43] and T2D2 [44, p. 2]. They are proprietary software developed by businesses and cannot be freely accessed. Their technology remains proprietary, closed source, and not accessible for obvious reasons; hence cannot be included or evaluated in this study.

Nevertheless, even if the code and the data are provided online to be accessible by interested stakeholders (e.g., construction and facility managers and building inspectors), those resources are still very limited in many ways. For example, (1) they require the user to be able to operate them using a command-line interface since many of the python scripts have to be run in the terminal, (2) they assume that the users can set up several things on their own – for instance, there are no instructions on these GitHub pages about how to set up the required machine learning libraries, except simply stating that they are required, e.g., Requirements – PyTorch, and (3) they do not generate a final report summarizing the cracks identified and the analyzed data, which is one of the most important information for construction and facility managers.

Many of the construction and facility managers that are not familiar with coding or do not interact with computers daily might not have the required technical literacy to operate the command line interface or figure out how to set up machine learning libraries on their own, and they would benefit by having a more user-friendly way to do the crack detection and analysis. To address that, ABECIS was developed as a freely accessible and user-friendly crack detection software with an intuitive Graphical User Interface (GUI). ABECIS is designed to work out-of-the-box, with minimal setup by the user. There is no need to go and download the pre-trained model and set up the software directory since it will perform the setup on its own. Moreover, once the input images are selected, the rest of the tasks, from the analysis to the report generation, are also done autonomously. For the parts of the cracks detected with uncertainty, the user will have a choice to intervene manually and reject the wrong identifications. Afterward, a qualitative and quantitative report will be generated along with annotated crack images.

### 2.3. Crack Detection Algorithms

Many algorithms currently exist for the detection of objects. However, almost all the algorithms can be generally classified into 4 broad categories. These algorithms can be classified into (a) Semantic Segmentation, (b) Classification and Localization, (c) Object Detection and (d) Instance Segmentation algorithms (Figure 1). Semantic Segmentation classifies every pixel in the image into classes (e.g., wall and crack). Classification and Localization algorithms only show the location of a single object. Object Detection algorithms are superior to the former two since they detect each object (e.g., in Figure 1) and its location, but not shape and size. Instance Segmentation is the most powerful technique among the four because it provides information about each object's location, shape and size. Therefore, for this study, instance segmentation was used as the main detection algorithm.

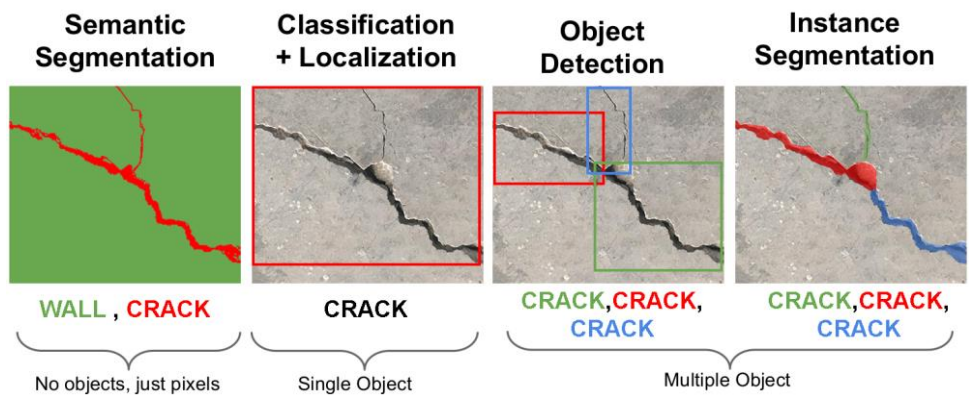

Figure 1: General Types of Object Detection Algorithms available at present

Hafiz et al. [45] did a survey on the state-of-the-art instance segmentation algorithms. Their study indicated that notable instance segmentation algorithms include PANet, YOLACT, and Mask R-CNN. PANet [46] and YOLACT [47] algorithms have their code on GitHub repositories but are no longer actively maintained and updated. PANet uses an adaptive feature pooling technique to funnel useful information in each feature level through the neural network for better mask prediction. YOLACT achieves real-time instance segmentation by breaking the instance segmentation into two subtasks. Nevertheless, non-updated code makes PANet and YOLACT non-ideal due to the deprecation of dependent python packages over time. Regarding Mask R-CNN, its official GitHub repository [48] states that Mask R-CNN has been deprecated and is now implemented into an instance segmentation framework called Detectron2 [49, p. 2], which is actively maintained by Meta Research. The comparison among different instance segmentation algorithms is beyond the scope of this study, and given the current state of Detectron2 regarding updateability and overall reported performance, the authors decided to use Detectron2 as the algorithm used to develop ABECIS.

## 3. Background

### 3.1. Data Collection Methods

The ABECIS system is designed in a very flexible way such that it can accept a variety of inputs produced by different methods of data collection – from very accessible, simplistic methods such as photos taken using smartphones to complex, professional methods such as photos taken using industrial drones such as the DJI Matrice 300 RTK (Figure 2), or autonomous robotic platforms as suggested by [50]. The only requirement is that the input must be an RGB image. For this study, two methods for image collection, namely cameras on (1) an industrial drone and (2) smartphones, were used.

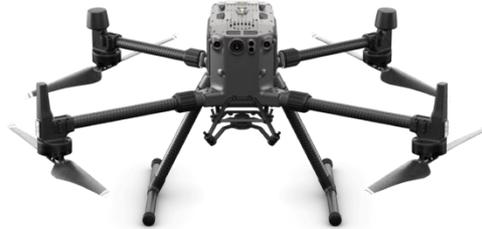

Figure 2: DJI Matrice 300 RTK [51]

Although the ABECIS system can function independently of the input method used, the quality of results and analysis produced may differ depending on the quality of input images from different methods. For instance, the images taken by the industrial drones will have more metadata that can be extracted for further analysis (e.g., the distance of the drone from the wall), which can be used to approximate the dimensions of the items of interest (in this case the length of a concrete crack). However, with the images taken by the smartphone, the data for the distance between the smartphone camera and the wall is missing; therefore, the dimensions of the crack cannot be approximated. Moreover, the difference in methods will also affect the accessible locations of the images for crack detection (e.g., drones can facilitate capturing images for crack detection in the façade of a 30[th]-floor building compared to a manual inspection using a smartphone).

### 3.2. Crack Classification Terminology

The classification of the concrete cracks for this study has been done in accordance with the ACI Concrete Terminology report by American Concrete Institute [52]. In particular, concrete cracks have been classified into 3 categories based on their orientation (Horizontal Crack, Vertical Crack and Diagonal Crack). Cracks that are either 180˚ or 90˚ to the horizontal are classified as horizontal and vertical cracks, respectively. Other cracks are classified as diagonal cracks (Figure 3). It is assumed that the images of the cracks are upright and not rotated, taken from a horizontal position, and with a camera orientation perpendicular to the plane being inspected.

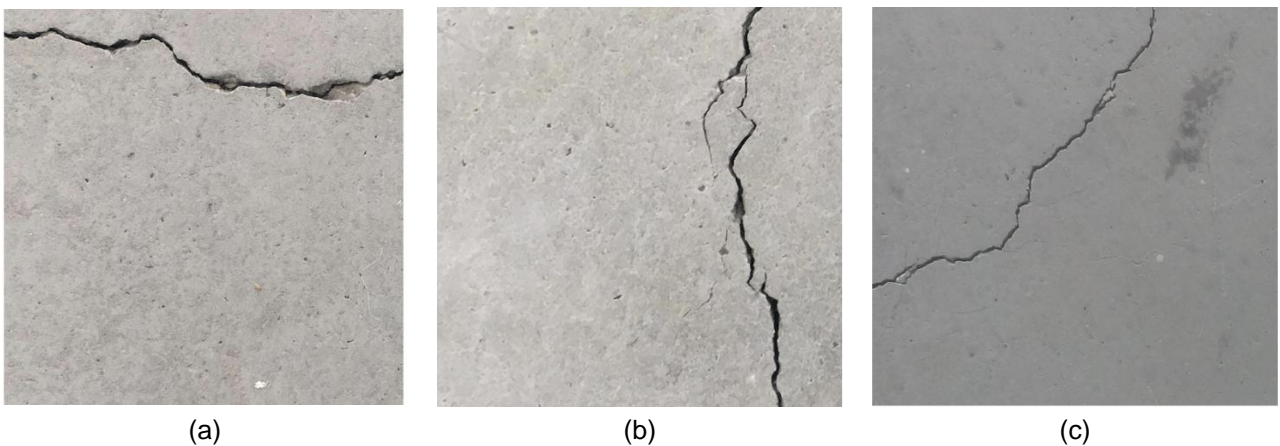

(a)  (b)  (c)

Figure 3: Different types of cracks classified in ABECIS: (a) horizontal, (b) vertical, and (c) diagonal from a horizontal perspective (source [53])

## 3.2. Deep Learning and Crack Detection

Deep Learning is a technique in Artificial Intelligence (AI) that uses artificial neural networks and is widely used to classify images [54]. Traditionally, crack detection algorithms use non-AI techniques such as Local Binary Patterns and shape-based algorithms [54]. Since 2012, deep learning algorithms have gained popularity and outperformed traditional detection methods [22].

### 3.2.1 Open-source deep learning frameworks

There has been a significant increase in publicly available open-source deep learning software libraries in recent years, allowing users to quickly create their custom deep learning models. For this study, a computer vision technique called instance segmentation - *the task of detecting the distinct objects of interest appearing in the images* – is used. Many popular instance segmentation algorithms already exist, such as PixelLib [55]. However, Detectron2 [49, p. 2], a state-of-the-art instance segmentation algorithm developed by Meta (formerly Facebook) Research, has been used for this study. Detectron2 is an improvement on Mask R-CNN and is updated more frequently than other algorithms [49, p. 2] [56].

### 3.2.2 Crack Detection Algorithm

The algorithm must first be trained to detect cracks using instance segmentation. The training requires many pre-labeled images. This dataset can be obtained by oneself or using publicly available datasets. Özgenel et al. [53] have publicly shared their dataset of concrete crack image data, which has been used in this research to train the model for concrete wall crack detection. For this study, the authors did a manual polygonal annotation of 800 images from [53] using LabelMe software [57] [58](e.g., Figure 4) so that those images could be used for instance segmentation training for model development. A sample of the labeled images developed for this study is publicly shared and available at [59].

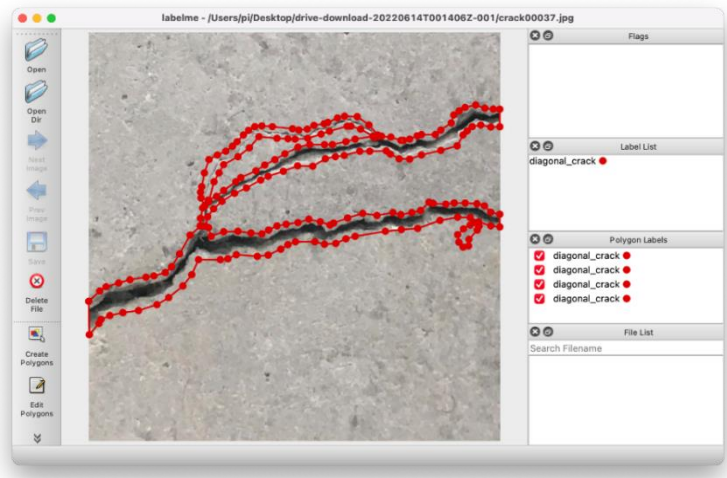

Figure 4: Polygonal annotation of images by the authors using LabelMe [57]. Labeled images used for model training.

## 4. Research Methodology

### 4.1. Real World Constraints and Assumptions

Numerous constraints and situations can interfere with the crack detection algorithm, producing false results in the real world. Image-processing results can be heavily affected depending on the camera angle, distance to the wall, shadows in images, and blind spots. For this study, the ABECIS system assumes the following from the images captured from the real world:

- The images are upright, not rotated and taken from a horizontal position.
- The images only contain the object of interest (wall) and no other objects.
- All the cracks, whether naturally occurring or intentionally man-made, will be detected (e.g., some architectural features might contain gaps or openings in the walls for aesthetic purposes)
- Camera orientation is perpendicular to the plane of the object under inspection.

These assumptions are made because unless the images taken by inspectors or drones are preprocessed before applying the crack detection algorithm, a high number of undesirable false-positive and false-negative errors are likely to occur [16]. However, preprocessing is currently not covered by the ABECIS system.

**4.2. System Architecture**

The overall system architecture for ABECIS is an expansion of the authors' work presented in [60]. It includes the human input and machine automation parts (Figure 5). Initially, the system will prompt the user to select a directory on their computer containing all the images to be analyzed for cracks. Afterward, the user will need to adjust the lower and upper confidence score threshold. Confidence scores are numbers between 0 and 100 used to describe how well the algorithm thinks it has detected a crack (100 being a perfect detection). The threshold values set a cutoff point such that identified cracks with a confidence score below the threshold value will be rejected. However, there may be some images of non-cracks for which the software is highly confident that they are cracks, resulting in false positives. In some scenarios, there might be objects which resemble cracks (e.g., dents or scratches on the walls, intentional man-made crack-like grooves, such as small gaps between two walls). For such cases, the algorithm might not be certain of the cracks identified and might require the assistance of a human operator to make a judgment (i.e., verification/intervention).

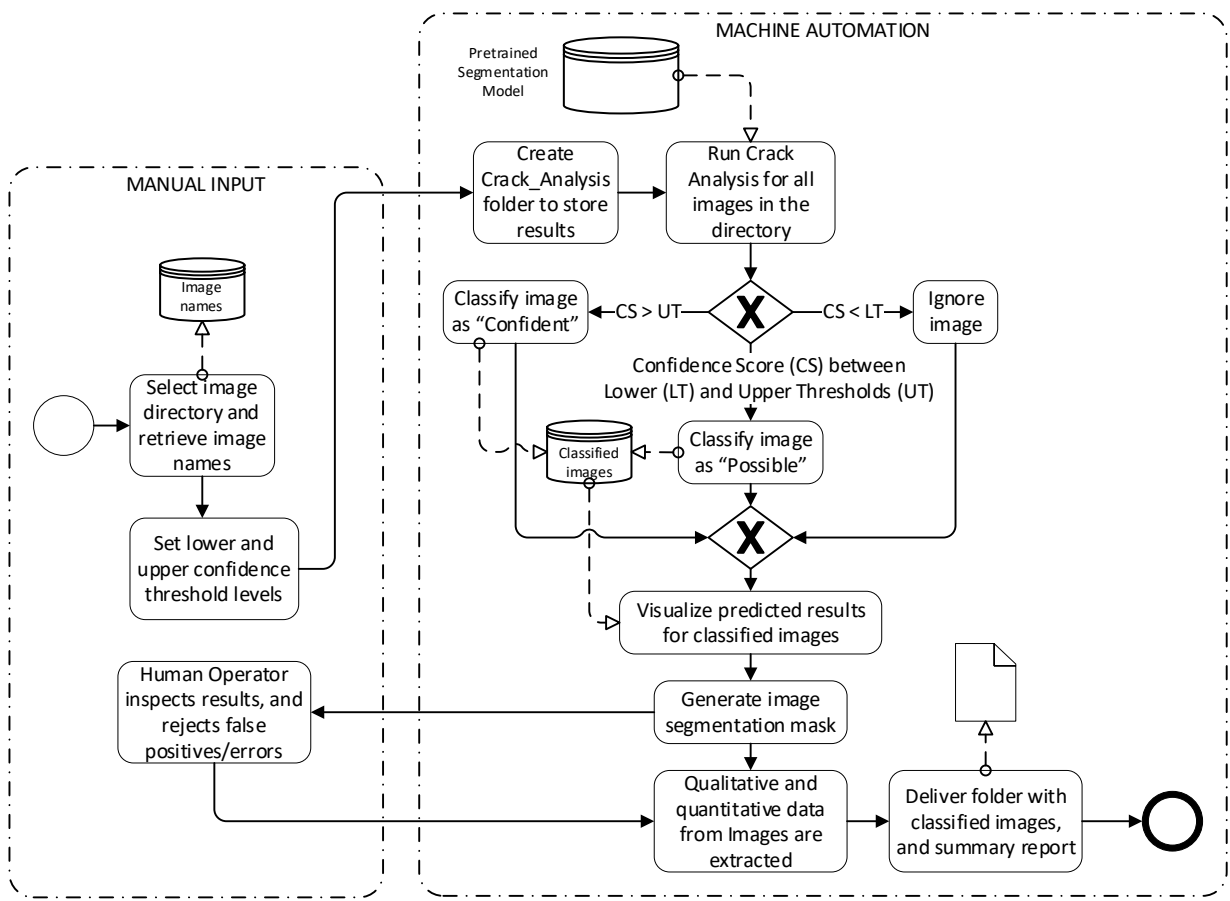

Figure 5: System architecture of the Automated Building Exterior Crack Inspection System (ABECIS) (expanded from [60])

Therefore, ABECIS has two threshold values (upper and lower) for the confidence scores, which work as shown in Figure 6. Any images containing crack detection confident scores higher than the upper threshold will be classified into a folder named "Confident," whereas identified cracks that fall within the upper and lower threshold will be classified into a folder named "Possible". Otherwise, the identified cracks are rejected. These two parameters are fully adjustable by the human operator and are to be decided by their judgment based on the environmental situation and difficulty of the crack detection task.

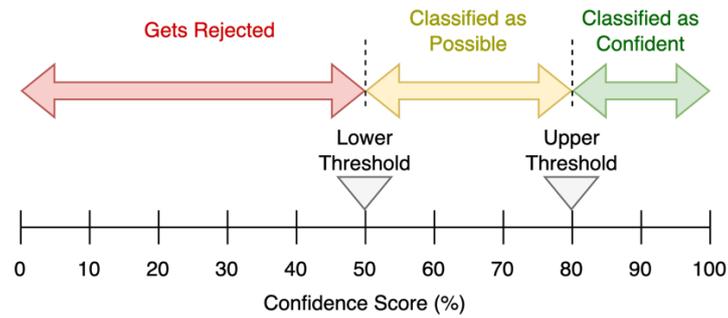

Figure 6: Confidence threshold in ABECIS

After the cracks are identified and classified into the "Possible" and "Confident" folders, the segmented masks of the images (black and white binary images where white represents the crack regions) are generated for quantitative analysis (e.g., length and area of the crack). The process is done completely autonomously by the software. Afterward, the human operator can use the Graphical User Interface (GUI) to inspect the crack detection results in the "Possible" and "Confident" folders and reject any false positives. Finally, taking this human input into account, the software can generate a detailed report for individual images with the types of cracks, including the time the image was taken (identified from metadata for monitoring purposes) and the length and area of the wall cracks in pixels and percentages. Therefore, the algorithm contains two instances where human intervention is required (initially to set the threshold parameters and finally to verify the outputs). Outside this, it is fully automated. It is worth mentioning that ABECIS could run completely autonomously with predefined thresholds. However, construction places are extremely dynamic environments and present a wide variety of conditions and scenarios. Allowing the human operator to provide some minimum input on the system allows for more robust and reliable results that can be used in real construction projects. ABECIS can be downloaded from the GitHub repository [5] and used by construction and facility managers and building inspectors with minimal setup using their own inspection images.

## 5. Deep Learning Model and Software development
### 5.1. Training the Instance Segmentation Neural Network

The image segmentation model was developed using 800 images and the Detectron2 library using the Özgenel et al. [53] dataset labeled by the authors. The training dataset consists of 700 images, and the testing dataset consists of 100 images. Example images of the training and testing datasets can be seen in Figure 7. The model was trained using COCO-Instance Segmentation, mask_rcnn_R_50_FPN_3x as the pre-trained model. Although Detectron2 contains more pre-trained models, such as C4 and DC5, FPN was chosen because it provides the best speed/accuracy tradeoff, according to Detectron2 documentation [61, p. 2].

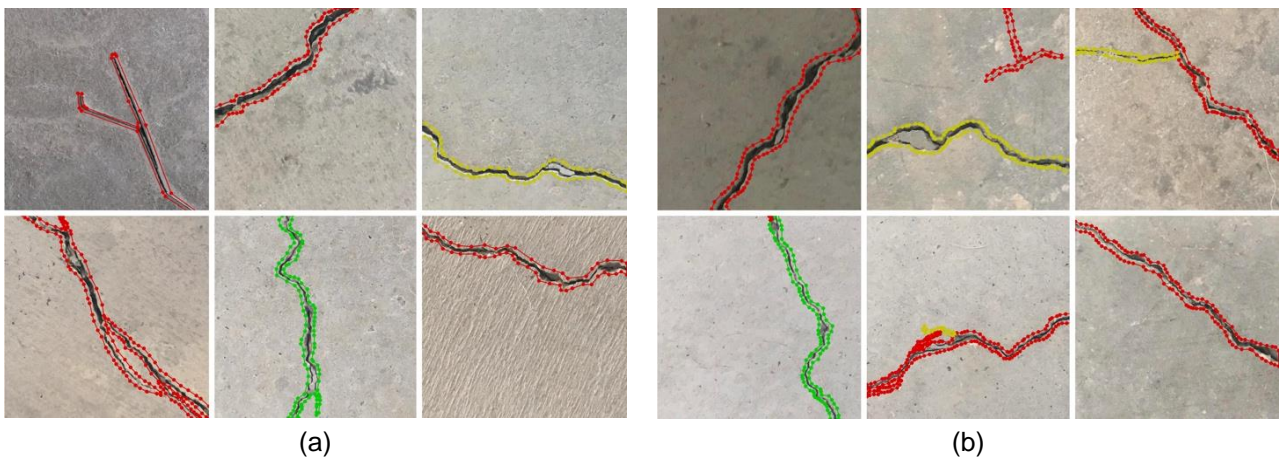

(a)  (b)

Figure 7: Examples of images used for the (a) training and (b) testing datasets (red = diagonal cracks, green = vertical cracks, yellow = horizontal cracks). The dataset was prepared via polygon labeling by the authors in LabelMe using raw images from [49, p. 2]. A subset of the labeled images developed for this study can be found in [59]

Cross-entropy loss [62] and accuracy [63] were calculated to measure how well the model performed. Cross-entropy loss of 0 indicates a perfect model, and an accuracy of 1 indicates a perfect model, too. After training for 3,000 epochs, the cross-entropy loss of the model reached close to 0.1 (Figure 8a), and the accuracy was 0.97 (Figure 8b). This indicates that the trained model is reliable.

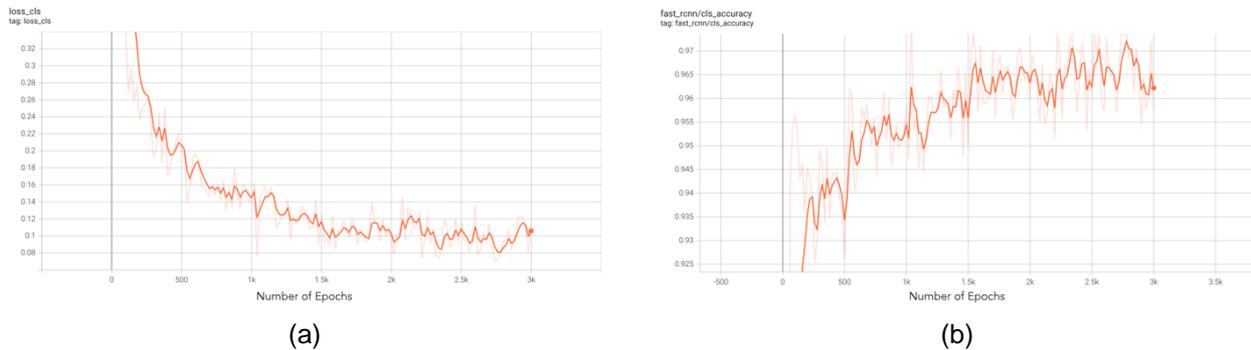

Figure 8: (a) Cross-Entropy Loss and (b) Accuracy during the training of the Neural Network model

**5.2. Software Features and Development**

Once the image segmentation model was trained, a cross-platform software application was developed using Detectron2 [49] for the deep learning component, PyQt6 [64] for the Graphical User Interface (GUI), and OpenCV for image processing [65]. The application has been tested on Windows and macOS. Although it has not been used in Linux, it should work in Linux, too. Full installation instructions and code is available in the GitHub repository [66]. All the functions described in Figure 5 have been implemented in the ABECIS Graphical User Interface (Figure 9).

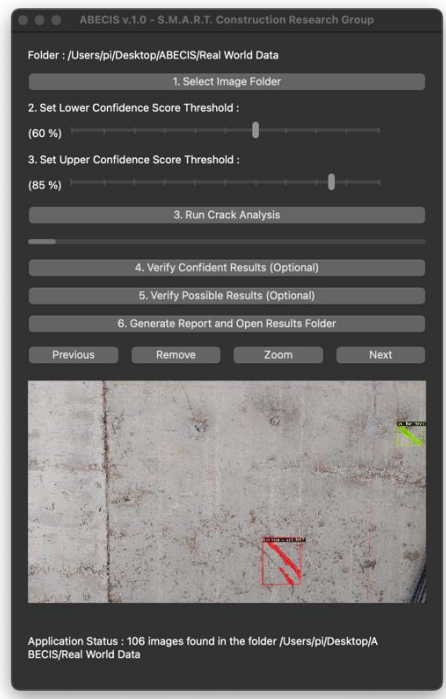

Figure 9: Graphical User Interface (GUI) for the ABECIS Software

The software uses multithreading; therefore, human operators can begin inspecting the results and start rejecting false positives without having to wait for the entire analysis process to finish.

**5.3. Crack Length Estimation**

Once the black and white binary masks of the crack detection results are generated, the length of the cracks can be approximated by converting the shape into a one-dimensional object, removing the width dimension using a morphological thinning operation [67]. This operation is an image processing method to reduce the shape of the object into a 1-pixel wide line, called the topological skeleton of the object. This operation is also handled by the ABECIS Graphical User Interface. The process is described by Equation 1.

$$A \wedge B = A - (As^*B) = A \cap (As^*B)^c \quad (1)$$

Where A is the set containing the image pixels with the value 1 (white), B is the structuring element, s* is a hit-or-miss transformation used in morphological image processing [68], and c is the complement.

As shown in Figure 10, the original image is analyzed for cracks, and the cracks are segmented. Then, the segmented cracks are converted to a binary mask, followed by the thinning process of the lines to 1-pixel width. Once the lines have a width of 1 pixel, the total length of the cracks in the image is simply the number of white pixels belonging to each crack. Without a metric reference in the image or information regarding the scale, it is not possible to get a proportional conversion between pixels and real-world measurements.

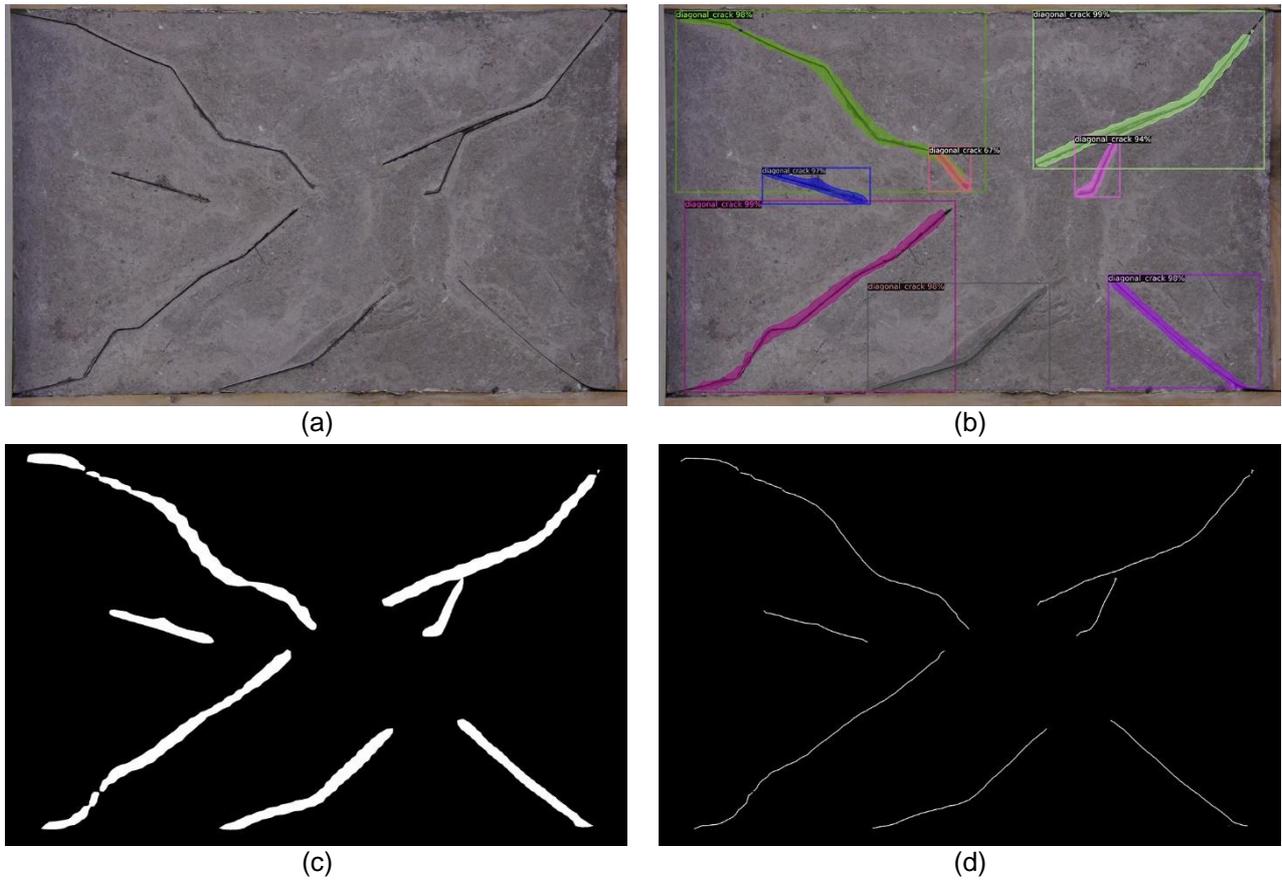

Figure 10: Different components of the crack length estimation process showing (a) original image, (b) instance segmentation of cracks, (c) binary mask resultant from the segmentation process, and (d) morphological thinning operation applied to the binary mask, resulting in lines with a width of 1-pixel.

## 6. Experimentation

Images taken in different environments using different methods were used to evaluate the performance of the ABECIS crack detection software. The results were evaluated (1) visually and (2) using the Intersection over Union (IoU), a metric to evaluate the performance of instance segmentation algorithms based on the extent of overlap of two boxes. The software has been tested in three different environments, (1) a controlled environment in the laboratory with images taken by a commercial drone, (2) a construction site with images taken by a smartphone camera, and (3) an outdoors environment around the New York University Abu Dhabi campus with images taken by a commercial drone (DJI Matrice 300 RTK).

## 6.1. Laboratory Experiment (Drone)

A movable mockup wall and a drone were used for the laboratory test (Figure 11). The mockup wall consisted of a concrete surface (approximate dimensions 63 cm height by 76 cm long) with a wood frame and 7 diagonal cracks. It was mounted on an aluminum frame with wheels for ease of transport. The drone was a DJI Matrice 300 RTK commercial. Altogether, 14 images were taken (the images can be found in [69]) and analyzed using ABECIS. The results (i.e., IoU and length) from these images are in Appendix B (Table 6).

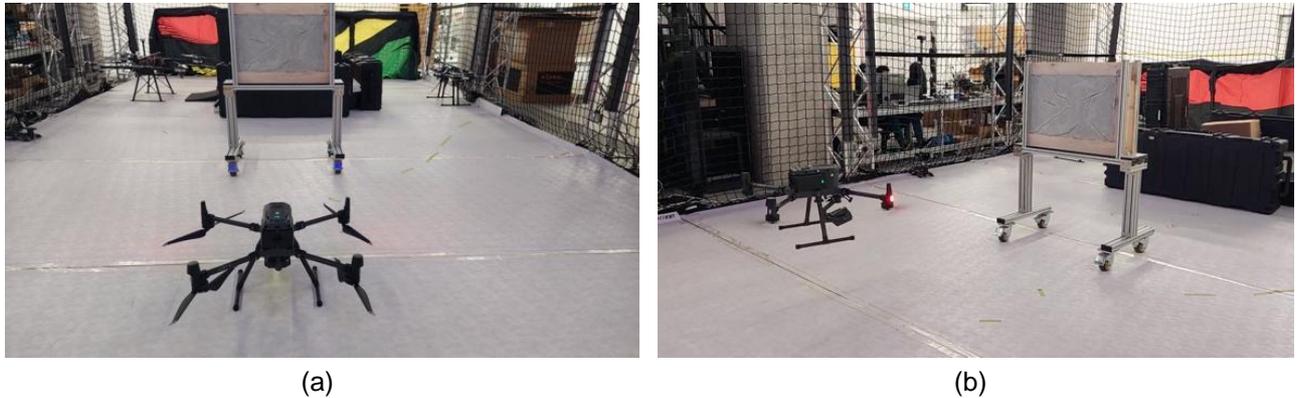

(a)                                                                                          (b)

Figure 11: General views of the experimental setup showing the mockup wall with concrete cracks and the drone used in (a) standby mode and (b) in operation (i.e., flying and taking images)

From the visual inspection of the different images and the result summarized in the report, it was found that the ABECIS system was able to successfully detect all the cracks in the controlled laboratory environment (Figure 12a), with just a small number of false positives. The false positives occurred when the image contained the surrounding objects in addition to the wall. In such cases, surrounding objects (similar to cracks) were classified as cracks (Figure 12b).

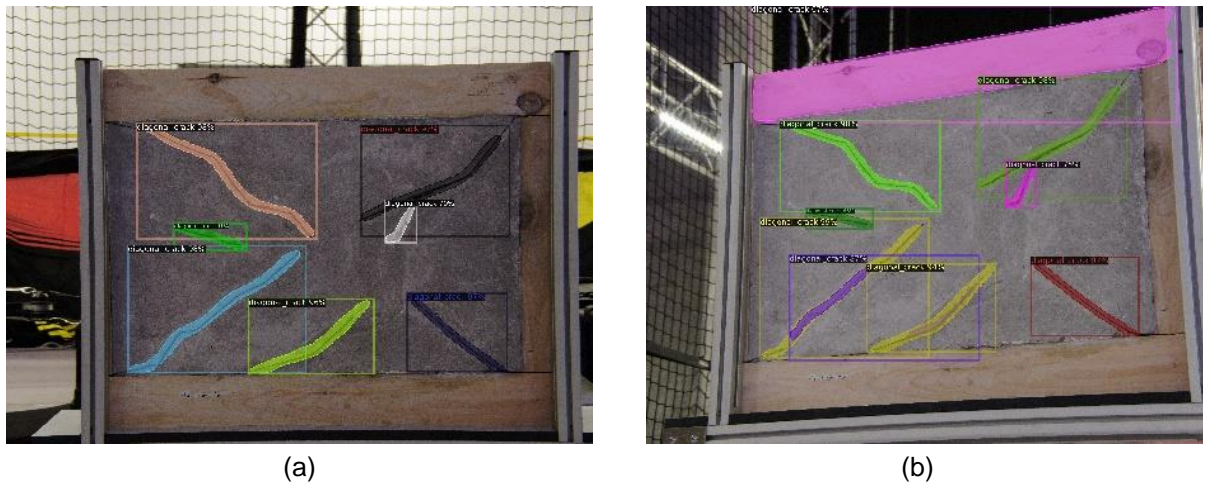

(a)                                                                                          (b)

Figure 12: Crack detection results from the experiment in the lab. (a) Annotations (b) False-positive segmentation of a non-crack object (upper member frame surrounding concrete wall).

## 6.2. On-Site Experiment (Smartphone Camera)

For the on-site test, the authors visited an ongoing construction site at the New York University Abu Dhabi campus and took images using an Android smartphone with a 1.9 Megapixel camera. Altogether 14 images were taken (the images can be found in [69]), and the reports were generated using ABECIS software. The results can be found in Appendix B (Table 7). Although there are not many cracks in the construction site, the ones present were correctly detected with varying degrees of accuracy (Figure 13).

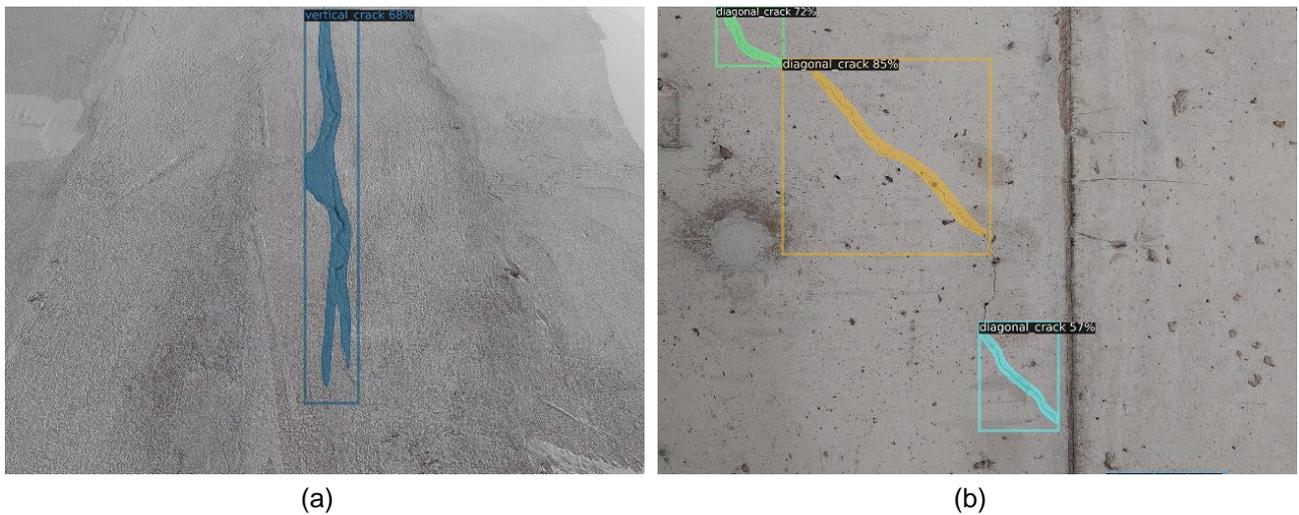

Figure 13: Examples of correctly detected (a) vertical and (b) diagonal cracks from the indoor construction site images taken using a smartphone

However, there are also several false positives. These are mainly due to objects that are very similar to cracks (e.g., scratches, markings on walls, pipes, wires). Some examples can be seen in Figure 14. It is worth mentioning that a real construction site is an extremely difficult scenario for a segmentation process. Nonetheless, all the true positives were detected, and all the resultant false positives can be manually discarded with an easy manual operation from the ABECIS GUI.

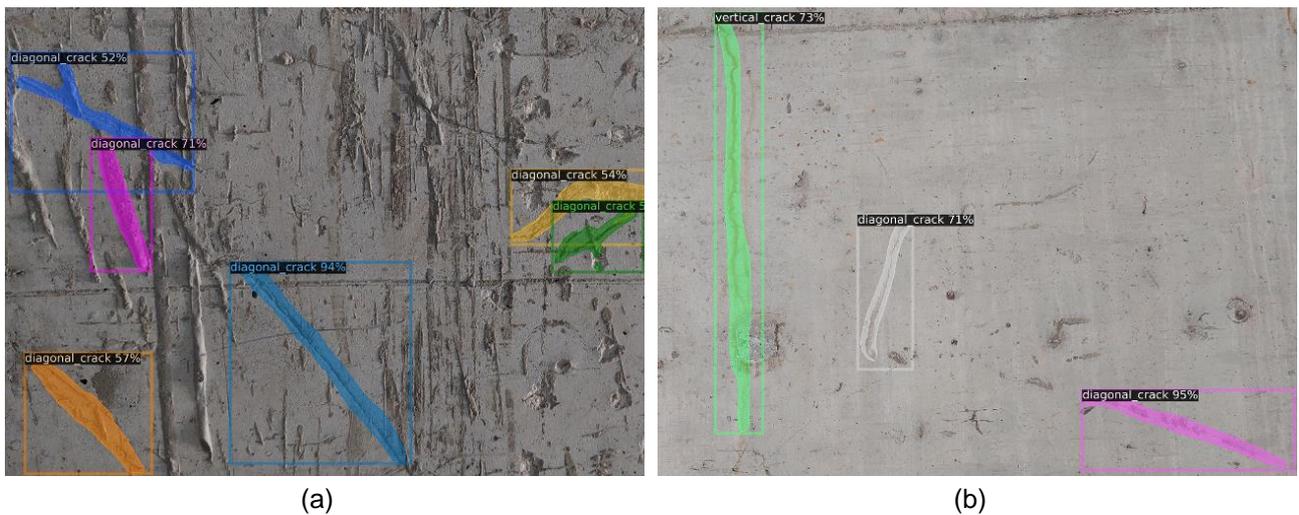

Figure 14: Construction Site false positive examples. (a) Scratches identified as cracks. (b) Liquid stains identified as cracks.

### 6.3 Outdoor Experiment (Drone)

Moreover, for the outdoor test, the authors also took images of the surrounding areas of the New York University Abu Dhabi campus using a DJI Matrice 300 RTK commercial drone. Altogether, 14 images were taken (the images can be found in [69]), and the results can be seen in Appendix B (Table 8). A few cracks found were detected (e.g., Figure 15). However, the algorithm occasionally missed very thin hairline-type cracks (Figure 15b).

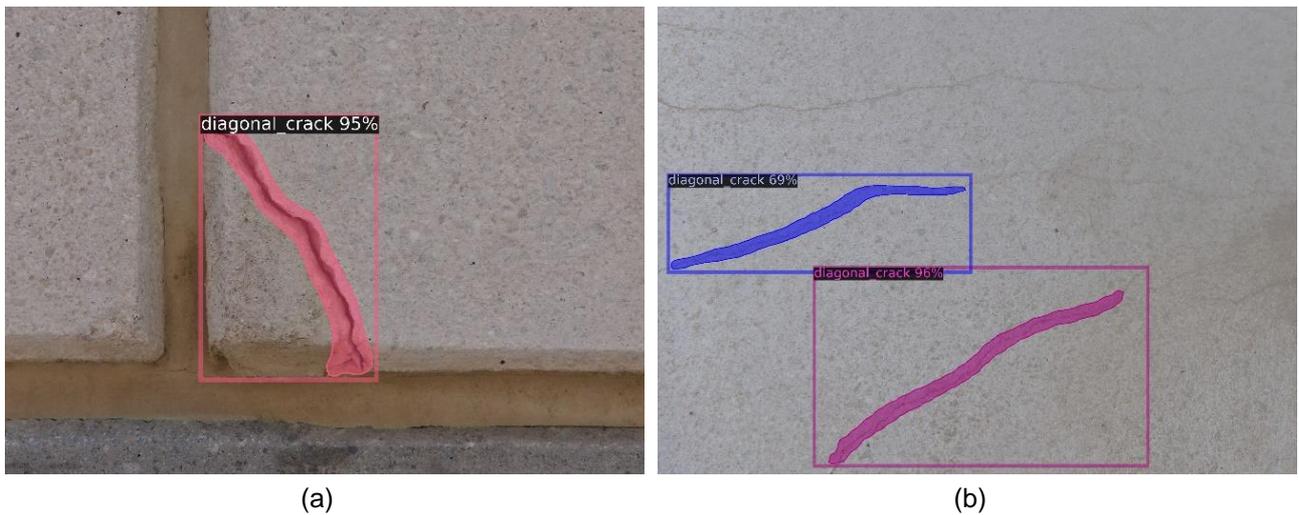

Figure 15: Examples of correctly detected cracks from outdoor images taken with a drone. (a) Single diagonal crack. (b) Multiple hairline cracks.

There were a few false positives in the outdoor drone experiment. Many were related to objects resembling cracks, such as the control joints with fillers or areas between the walls or concrete slabs, as in Figure 16a or the gap between two concrete slabs, as in Figure 16b.

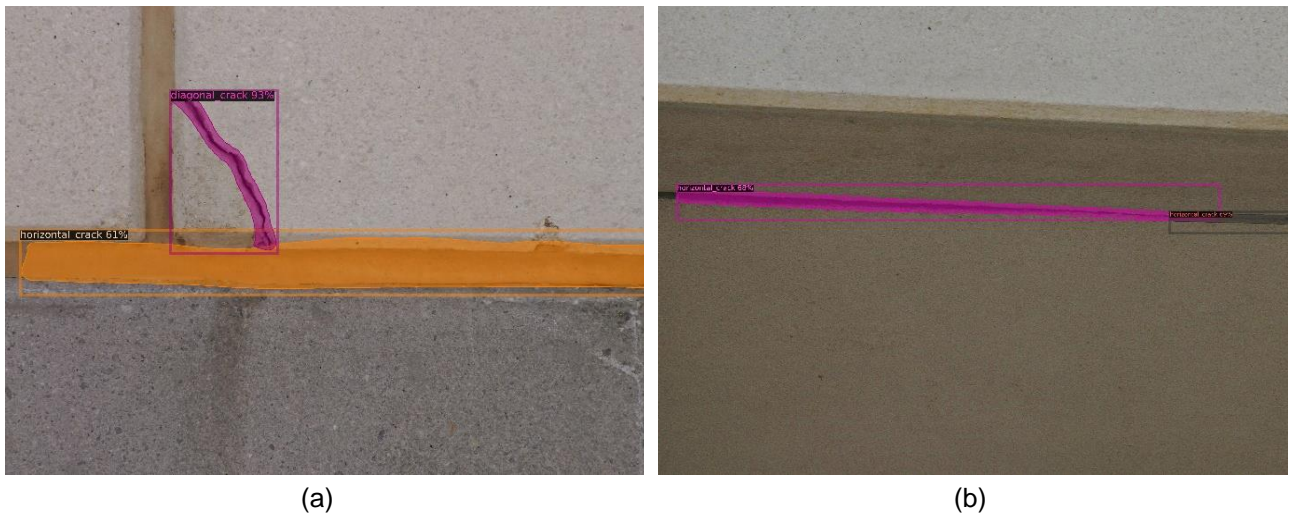

Figure 16: False positives examples from the outdoor experiment. (a) Control joint/filler between walls, identified as a horizontal crack. (b) Area between parapet wall (vertical element) and ceiling panel (horizontal element) detected as crack.

## 7. Results and limitations

All the results are provided in Appendix B and C. The Repository can be found with detailed information at https://github.com/SMART-NYUAD/ABECIS-Results.

### 7.1. Intersection over Union (IoU) Evaluation

For the evaluation of the performance of the ABECIS system, Intersection Over Union (IoU), a method to quantify the percent overlap between the target mask and the obtained output, was calculated for the 14 images in three categories (i.e., Lab, On-Site and Outdoors). An IoU score of 1 indicates that the model performs perfectly, whereas a score of 0 indicates that the prediction and the ground truth do not overlap at all. The IoU was calculated for a total of 42 (14x3) images in three experiments, and the summary statistics are shown in Table 2 and plotted as Box and Whisker Plot in Figure 17.

To account for the effect of human verification/intervention, two different types of summary statistics for the IoU were calculated for each of the three experiments. One with all the cases and another with True Positives only (indicating the scenario where the inspector reviewing the output would reject those cases). IoU (All) refers to the IoU statistics calculated from the raw crack analysis data directly obtained from ABECIS software, whereas IoU (True Positives) refers to the IoU statistics after the human operator has removed all false positives from ABECIS GUI.

Table 2. Summary statistics for the Intersection over Union of the three experiments

| Experiment | Lower Adjacent | Lower Quartile | Median | Upper Quartile | Upper Adjacent | Outliers |
|---|---|---|---|---|---|---|
| Indoor – Lab (Drone) - All | 0.505 | 0.642 | 0.686 | 0.853 | 0.999 | N/A |
| Indoor – Lab (Drone) - True Positives | 0.642 | 0.678 | 0.838 | 0.938 | 0.999 | N/A |
| On-Site – Construction Site (Smartphone) - All | 0.000 | 0.000 | 0.186 | 0.903 | 0.968 | N/A |
| On-Site – Construction Site (Smartphone) - True Positives | 0.839 | 0.889 | 0.967 | 0.971 | 0.981 | N/A |
| Outdoor – Campus (Drone) - All | 0.370 | 0.550 | 0.958 | 0.968 | 0.972 | N/A |
| Outdoor – Campus (Drone) - True Positives | 0.370 | 0.550 | 0.958 | 0.968 | 0.972 | N/A |

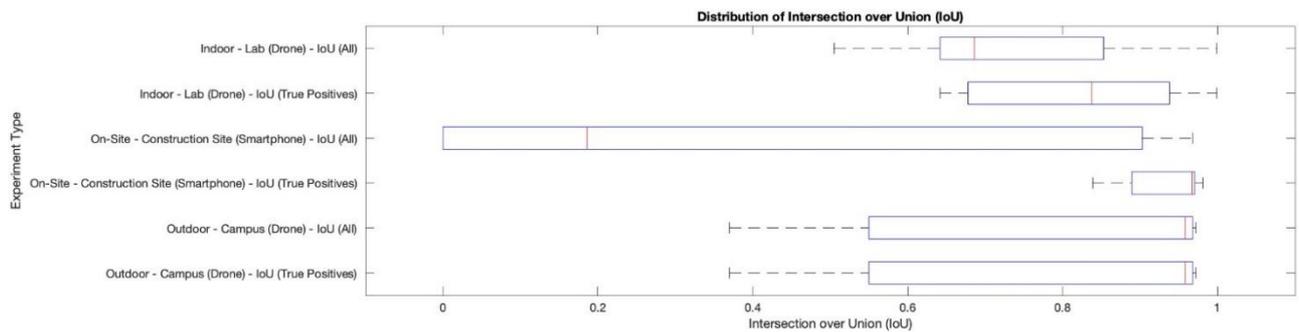

Figure 17: Distribution of IoU for different experiment types. IoU (All) refers to the distribution of IoU from the raw predictions from the algorithm. IoU (True Positives) refers to the distribution of IoU after the human operator has removed all false positives from the results.

For the IoU for all images, the median of IoU is (1) 0.686 for the indoor crack detection experiment in a controlled lab environment using a commercial drone, (2) 0.186 for the on-site crack detection at a construction site using a smartphone and (3) 0.968 for the outdoor crack detection at the NYUAD campus using a commercial drone. The interquartile range is the smallest for the Indoor Lab experiment, followed by the Outdoor Campus Experiment and the On-Site Construction Site Experiment.

However, only considering the true positives after human verification, the median of IoU is (1) 0.838 for the indoor crack detection experiment in a controlled lab environment using a commercial drone, (2) 0.967 for the on-site crack detection at a construction site using a smartphone and (3) 0.968 for the outdoor crack detection at the NYUAD campus using a commercial drone. The interquartile range is the smallest for the On-Site Construction Site Experiment, followed by the Indoor Lab experiment and the Outdoor Campus Experiment.

These results indicate that with little or no human verification, ABECIS worked best for the Outdoor Drone inspection, even when compared to the controlled lab experiment. This can be attributed to the condition of the exterior walls not being covered with any obstacles and having a relatively constant background color, making them good candidates for object detection. Moreover, clear weather and wall paint of light color also played a role in the successful detection of the cracks, increasing the contrast between the crack and non-crack pixels.

The results from the indoor controlled lab environment with the drone show a median IoU smaller than the outdoor campus experiment. This can be attributed to the number of false positives (e.g., wood planks in the mockup wall and noisy background environment) and unfavorable lighting conditions. Lastly, the construction site experiment has a very large interquartile range and the lowest median. This means that ABECIS had the lowest performance on an active construction site. This is because the construction site is unstructured and contains many temporary elements, such as exposed wires and scaffolding, which were falsely detected as cracks.

However, considering that all the IoU median values of true positives are greater than 0.8, it can be argued that ABECIS works very well in all indoor, outdoor and construction site environments, using both smartphone and drone cameras when the machine learning algorithm is aided by the human verification process. The IoU score of the trained model can be improved by retraining the model with the most common falsely detected objects (e.g., wires) and removing them from the results. However, this is impractical since many kinds of objects can cause false positives.

### 7.2. Crack length estimation

Another of the features provided by ABECIS is the crack length estimation. It is worth mentioning that the length measurement given in metric units is only possible if the image has a scale reference to obtain a proportional relationship between the pixel and real-world dimensions or if the image was taken with a device able to measure the distance to the wall (such as the DJI Matrice 300 RTK used in this experiment). The crack length is a useful metric that can assist inspectors in getting a better assessment of the state of the building and if the crack needs urgent repairing or not (also, it would be useful information to generate preliminary repair scopes and cost estimates); therefore, it is a valuable asset to be included in the final report generated by ABECIS. In order to test the effectiveness and robustness of the estimations calculated by ABECIS, the controlled lab environment with the mockup wall was used. The ground-truth measurements and the predicted lengths of the cracks present in the mockup wall can be seen in Figure 18.

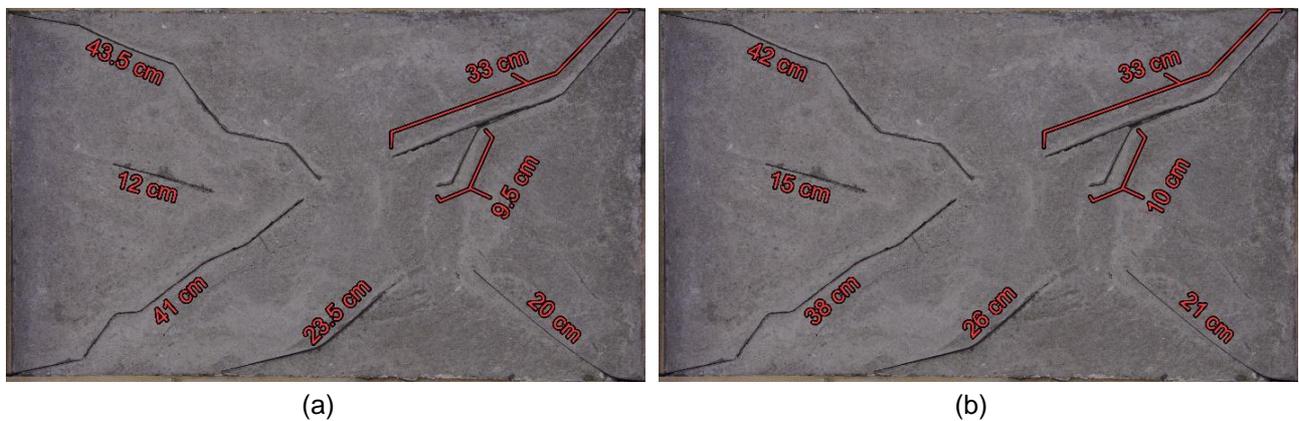

(a) (b)

Figure 18: (a) Original dimensions of cracks on the mockup wall, and (b) corresponding predictions

The length estimations computed for each of the 14 images can be seen in Appendix C (Table 9). The results in the Appendix correspond to the sum of the estimated lengths for all the successfully detected cracks on each image. The computed error is the difference between this value and the sum of all the ground-truth distances. Table 3 and Figure 19 summarize the measurements from the 14 images. The error median is 8.2%, which is an acceptable margin to provide qualitative assessments of the crack lengths.

Table 3. Summary statistics for the % Error in Crack Length Estimation considering true positive cases

|  | Lower Adjacent | Lower Quartile | Median | Upper Quartile | Upper Adjacent | Outliers |
|---|---|---|---|---|---|---|
| % Error | 0.3 | 1.9 | 8.2 | 14.5 | 24.9 | N/A |

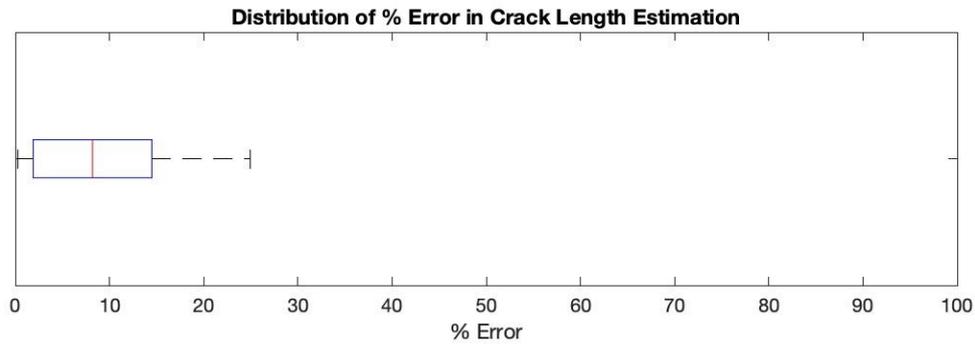

Figure 19: Distribution of % Error in Crack Length Estimation

### 7.3. Report generation

After the crack detection is completed and all the measurements have been computed, the results are automatically gathered into a comma-separated values (csv) file format. The generated report contains information such as the image filename, type of cracks detected, date and time extracted from the image metadata, the types of cracks present in the image, and some metrics such as the Number of Confident Cracks and their Average Confidence Score, as well as the number of Possible Cracks and their Confidence Scores. Moreover, the total crack length in pixels is also provided, and from here, the estimated total crack length in centimeters can be calculated provided that a reference object with known length is available in the photo (like in the case of the images taken using the drone. For the images taken using a smartphone, a reference should be added when taking each photo). An excerpt of a report generated is shown in Table 4.

Table 4. Excerpt of a report generated by ABECIS

| Filename | Date/Time Taken | Crack Types | No. of Confident Cracks | Average Confidence Score for Confident Cracks | No. of Possible Cracks | Average Confidence Score for Possible Cracks | Total Crack Length (pixels) | Estimate Total Length (cm) |
|---|---|---|---|---|---|---|---|---|
| lab_00003.jpg | 2022:03:31 16:21:40 | 'Diagonal Crack' | 3 | 93.6 | 2 | 82.0 | 185 | 164 |
| lab_00007.jpg | 2022:03:31 16:19:44 | 'Diagonal Crack' | 4 | 92.0 | 1 | 84.4 | 169 | 194 |
| lab_00011.jpg | 2022:03:31 16:22:06 | 'Diagonal Crack', 'Horizontal Crack' | 2 | 94.5 | 5 | 72.6 | 150 | 290 |
| lab_00006.jpg | 2022:03:31 16:22:06 | 'Diagonal Crack' | 4 | 97.4 | 4 | 77.4 | 516 | 364 |
| lab_00002.jpg | 2022:03:31 16:21:10 | 'Diagonal Crack' | 7 | 95.6 | 2 | 63.2 | 727 | 202 |
| lab_00012.jpg | 2022:03:31 16:20:29 | 'Diagonal Crack' | 6 | 97.1 | 2 | 72.5 | 609 | 209 |
| lab_00001.jpg | 2022:03:31 16:19:25 | 'Diagonal Crack' | 6 | 96.5 | 1 | 70.4 | 595 | 207 |
| lab_00008.jpg | 2022:03:31 16:19:58 | 'Diagonal Crack' | 6 | 96.3 | 1 | 83.8 | 611 | 186 |
| ⋮ | ⋮ | ⋮ | ⋮ | ⋮ | ⋮ | ⋮ | ⋮ | ⋮ |

**7.4. Limitations**

One of the limitations of ABECIS is the impact of cluttered and unstructured environments. Images with a cluttered and non-uniform background present a specially challenging scenario when it comes to object detection, making the performance more susceptible to false positives in the detection process. Nonetheless, it is expected that the main use of ABECIS would be in the maintenance stage of the lifecycle of a building, inspecting exterior façades that present a clear and uniform background for the most part. From the results obtained, ABECIS performs best under those conditions.

Another limitation is that estimating the length of the cracks identified (i.e., providing measurements in metric units) is only available if the image contains enough data (or a reference) to establish a proportional relation between pixel and real-world distances. That means that for pictures taken with a smartphone, a metric scale needs to be present in every image (which in real conditions might not be practical). However, most modern smartphone models have multiple cameras and even small LiDARs that would allow the software to extract enough metadata from the image to compute the required reference to estimate the length of the detected cracks.

The current implementation has only been tested in concrete-like façades. Although concrete is still the most used material in construction, modern buildings include different types of materials on their exterior façades, which could lead to different results in the segmentation process due to the different nature of the material. Other issues (beyond cracks) also manifest in the exterior of buildings (e.g., water stains/infiltration, sealant deterioration, thermal leaks, etc.), which can accelerate the deterioration of the building and compromise the efficiency of the different building systems. Although ABECIS only focuses on cracks, the algorithm can easily be expanded to identify other issues.

Finally, Detectron2 has been the algorithm selected for ABECIS. However, other instance segmentation algorithms could provide different performance or advantages. It was beyond the scope of this study to do a comparison/evaluation of different algorithms, but this could be considered in future studies.

**8. Conclusion and Outlook**

This study proposes an open-source Automated Building Exterior Crack Inspection Software (ABECIS) for construction and facility managers and building inspectors. The software uses an instance segmentation Artificial Intelligence model developed and trained by the authors using a custom dataset. The system was tested and evaluated in a laboratory (i.e., a controlled environment) and real-world scenarios using different ways to collect images (e.g., cameras on a commercial drone and a smartphone).

ABECIS allows the operation to verify the output provided (i.e., human-in-the-loop). From the raw output of the algorithm without the human verification, the median IoU is greatest for the outdoor crack detection experiments using a drone, followed by the indoor crack detection experiment in a controlled lab environment using a drone and lastly, for the indoor crack detection at a construction site using a smartphone. This indicates that ABECIS performs best for outdoor drone images with minimal human verification. Very often, false positives arise when the environment has too many obstacles and other objects in addition to the wall to be analyzed.

However, these IoU results improve significantly (to over 0.8) when a human operator selectively removes the false positives through the ABECIS user interface. Therefore, combining the ABECIS' predictions and human verification can offer very accurate crack detection for all cases (indoor and outdoor scenarios).

Interested readers can easily install and use our method by visiting the GitHub repository of ABECIS [5].

Ongoing work by the authors includes improving the quality of crack detection of ABECIS software by using a larger training dataset and obstacle detection systems to reduce false positives.

**Acknowledgment**

This work was partially supported by the NYUAD Center for Interacting Urban Networks (CITIES), funded by Tamkeen under the NYUAD Research Institute Award CG001. The authors would like to thank Eyob Mengiste from the S.M.A.R.T. Construction Research Group at NYUAD, Oraib Al-Ketan, Jinumon Govindan and Vijay Dhanvi from NYUAD CNC Machinist Machine Shop for their assistance with the manufacturing of the mockup wall for the experiment, and Nikolaos Giakoumidis from "Kinesis CTP" (Core Technology Platform) for his support with the drone during the lab experiment. Thanks also to William Fulton and Mohamed Rasheed from the Campus Planning and Projects Office at NYUAD for their continuous support and for allowing access to ongoing projects on campus.

**Appendix A**

The descriptions of categories in Table 5 are summarized as follows:

- Column A: Reference ID.
- Column B: The year the paper was published.
- Column C: Indicates if the technique proposed uses AI or traditional image processing (such as Canny Edge Detection [70]).
- Column D: The technique used to collect images for processing. UAV indicates a drone with a camera to collect images. Camera indicates a camera other than the one in the UAV.
- Column E: The type of structure (e.g., buildings, bridges) and/or the material the surface is made of (e.g., concrete, steel).
- Column F: The type of defect/item detected (e.g., cracks, absence of paint).
- Column G: Indicates if the paper provides a qualitative evaluation of the method proposed.
- Column H: Degree of autonomy of the proposed method. FA means fully autonomous, and no human intervention is required for the system to function. SA means semi-autonomous, and a significant amount of manual labor (e.g., manually collecting the input data, tweaking thresholds during the processing) is required for the proposed method. M means a fully manual approach with no level of automation involved.

Table 5. Summary of different papers identified during the literature review (in reverse chronological order)

| A | B | C | D | E | F | G | H |
|---|---|---|---|---|---|---|---|
| [11] | 2021 | ✓ | DSLR*, UAV | Concrete Bridge | Cracks | ✓ | SA |
| [25] | 2021 |   | UAV | Concrete Bridge | Cracks |   | SA |
| [23] | 2021 | ✓ | UAV | Concrete | Cracks | ✓ | SA |
| [13] | 2020 | ✓ | Smartphone | Asphalt | Cracks | ✓ | SA |
| [32] | 2020 |   | UAV | Metal Bridge | Paint Absence | ✓ | SA |
| [15] | 2020 | ✓ | UAV | Asphalt | Cracks | ✓ | SA |
| [8] | 2020 | ✓ | UAV | Concrete Water Tank | Cracks |   | SA |
| [7] | 2020 | ✓ | UAV | Concrete Bridge | Cracks | ✓ | SA |
| [36] | 2020 | ✓ | UAV | Mine Slope | Cracks | ✓ | FA |
| [26] | 2020 |   | UAV | Concrete Bridge | Cracks | ✓ | SA |
| [18] | 2020 | ✓ | N/A | Damage after earthquake | Cracks, spalling, severe damage | ✓ | SA |
| [19] | 2020 | ✓ | UAV | Concrete | Cracks | ✓ | SA |
| [20] | 2020 | ✓ | UAV | Asphalt | Potholes and cracks | ✓ | SA |
| [29] | 2020 |   | UAV | Concrete Bridges | Biological colonies, efflorescence, cracks and exposed rebar |   | SA |
| [30] | 2020 |   | UAV | Concrete Bridge | Cracks | ✓ | SA |
| [71] | 2020 | ✓ | UAV | Façades | De-blurring images for crack detection | ✓ | SA |
| [31] | 2020 | | UAV | Concrete Bridges | Cracks | | SA |
| [33] | 2019 | | UAV | Brick Building | Change in appearance | ✓ | SA |
| [16] | 2019 | ✓ | UAV | Metal Structures | Weld Defects | ✓ | SA |
| [72] | 2019 | | UAV | Concrete | Cracks | ✓ | SA |
| [73] | 2019 | ✓ | Climbing robot | Aircraft surface | Corrosion, stains, and cracks | ✓ | SA |
| [34] | 2018 | ✓ | UAV | Concrete | Cracks | | FA |
| [35] | 2018 | ✓ | UAV | Concrete | Cracks | ✓ | FA |
| [74] | 2018 | ✓ | N/A | Concrete | Cracks | ✓ | SA |
| [27] | 2018 | | UAV | Concrete Bridge | Cracks | | SA |
| [28] | 2018 | | UAV | Wind turbine | Cracks | | SA |
| [75] | 2018 | ✓ | UAV | Concrete Lab test | Cracks | | FA |
| [76] | 2018 | | UAV | Steel | Fatigue cracks | | M |
| [9] | 2018 | ✓ | UAV | Concrete Bridges | Cracks | | SA |
| [77] | 2017 | | UAV | Concrete Bridge | Cracks | ✓ | SA |
| [24] | 2017 | | UAV | Concrete Bridge | Cracks | ✓ | SA |
| [12] | 2017 | ✓ | UAV | Concrete Bridge | Cracks | ✓ | SA |
| [78] | 2017 | | UAV | Concrete Bridge | Cracks | ✓ | SA |
| [79] | 2017 | | UAV | Concrete Bridges & Tunnel | N/A | ✓ | FA |
| [80] | 2017 | ✓ | UAV | Pavement | Cracks | ✓ | SA |

*\* Digital single-lens reflex camera*

Main elements from Table 5 are broken down in Figure 20.

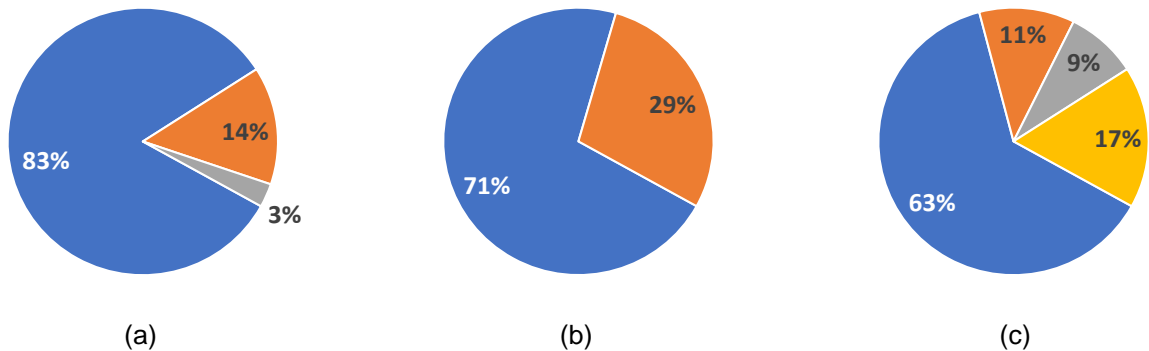

(a)          (b)          (c)

Figure 20. Graphical representation of the most representative columns from Table 5. (a) Level of automation – blue, orange, and grey for SA-semi-autonomous, FA-fully autonomous and M-manual, respectively. (b) Percentage of papers that provide a quantitative evaluation – blue yes, orange no. (c) Classification according to the type of structure – blue, orange, grey and yellow for concrete, asphalt, metal, and other, respectively.

## Appendix B

Processed data and images mentioned in the tables below can be accessed at [69].

Table 6. Intersection over Union Scores of Laboratory Drone Images

| No | Image Name | Crack Type(s) | Image (segmentation mask) | True Positives | False Positives | IoU (All) | IoU(True Positives) |
|---|---|---|---|---|---|---|---|
| 1 | lab_00017.jpg | 'Diagonal Crack' | 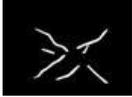 | 8 | 0 | 0.999 | 0.999 |
| 2 | lab_00001.jpg | 'Diagonal Crack' | 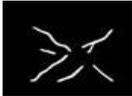 | 7 | 0 | 0.962 | 0.962 |
| 3 | lab_00003.jpg | 'Diagonal Crack' | 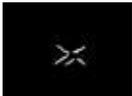 | 5 | 0 | 0.929 | 0.929 |
| 4 | lab_00004.jpg | 'Diagonal Crack', 'Horizontal Crack' | 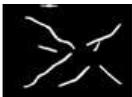 | 7 | 1 | 0.853 | 0.892 |
| 5 | lab_00002.jpg | 'Diagonal Crack' | 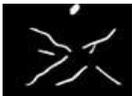 | 9 | 1 | 0.804 | 0.903 |
| 6 | lab_00008.jpg | 'Diagonal Crack' | 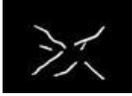 | 7 | 0 | 0.773 | 0.773 |
| 7 | lab_00012.jpg | 'Diagonal Crack' | 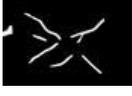 | 7 | 1 | 0.694 | 0.783 |
| 8 | lab_00018.jpg | 'Diagonal Crack' | 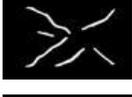 | 8 | 0 | 0.678 | 0.678 |
| 9 | lab_00006.jpg | 'Diagonal Crack' | 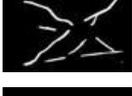 | 7 | 1 | 0.665 | 0.680 |
| 10 | lab_00022.jpg | 'Diagonal Crack' | 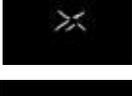 | 5 | 0 | 0.642 | 0.642 |
| 11 | lab_00007.jpg | 'Diagonal Crack' | 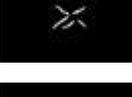 | 5 | 0 | 0.642 | 0.642 |
| 12 | lab_00011.jpg | 'Diagonal Crack', 'Horizontal Crack' | 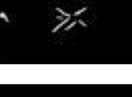 | 5 | 1 | 0.554 | 0.654 |
| 13 | lab_00019.jpg | 'Diagonal Crack' | 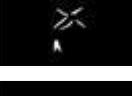 | 5 | 1 | 0.509 | 0.971 |
| 14 | lab_00021.jpg | 'Diagonal Crack' | 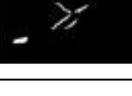 | 4 | 1 | 0.505 | 0.938 |

Table 7. Intersection over Union Scores of Outdoor Construction Site Images

| No | Image Name | Crack Type(s) | Image (segmentation mask) | True Positives | False Positives | IoU (All) | IoU (True Positives) |
|---|---|---|---|---|---|---|---|
| 1 | outdoor_site_00197.jpg | 'Vertical Crack' | | 1 | 0 | 0.968 | 0.968 |
| 2 | outdoor_site_00200.jpg | 'Vertical Crack' | | 1 | 0 | 0.967 | 0.967 |
| 3 | outdoor_site_00187.jpg | 'Vertical Crack' | | 1 | 0 | 0.946 | 0.946 |
| 4 | outdoor_site_00201.jpg | 'Horizontal Crack' | | 1 | 0 | 0.903 | 0.903 |
| 5 | outdoor_site_00168.jpg | 'Diagonal Crack' | | 2 | 0 | 0.848 | 0.848 |
| 6 | outdoor_site_00169.jpg | 'Diagonal Crack' | | 3 | 0 | 0.839 | 0.839 |
| 7 | outdoor_site_00003.jpeg | 'Diagonal Crack', 'Horizontal Crack', 'Vertical Crack' | | 1 | 3 | 0.252 | 0.981 |
| 8 | outdoor_site_00001.jpeg | 'Horizontal Crack', 'Vertical Crack' | | 2 | 3 | 0.120 | 0.976 |
| 9 | outdoor_site_00002.jpeg | 'Diagonal Crack' | | 1 | 3 | 0.089 | 0.969 |
| 10 | outdoor_site_00178.jpg | 'Diagonal Crack', 'Vertical Crack' | | 0 | 3 | 0.000 | N/A |
| 11 | outdoor_site_00136.jpg | 'Diagonal Crack', 'Horizontal Crack' | | 0 | 2 | 0.000 | N/A |
| 12 | outdoor_site_00174.jpg | 'Vertical Crack' | | 0 | 1 | 0.000 | N/A |
| 13 | outdoor_site_00195.jpg | 'Diagonal Crack' | | 0 | 6 | 0.000 | N/A |
| 14 | outdoor_site_00177.jpg | 'Diagonal Crack' | | 0 | 1 | 0.000 | N/A |

Table 8. Intersection over Union Scores of Outdoor – Campus - Drone Images

| No | Image Name | Crack Type(s) | Image (segmentation mask) | True Positives | False Positives | IoU (All) | IoU (True Positives) |
|---|---|---|---|---|---|---|---|
| 1 | outdoor_drone_00006.jpg | 'Diagonal Crack' | 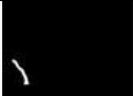 | 1 | 0 | 0.972 | 0.972 |
| 2 | outdoor_drone_00008.jpg | 'Diagonal Crack' | 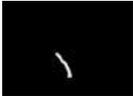 | 1 | 0 | 0.970 | 0.970 |
| 3 | outdoor_drone_00016.jpg | 'Diagonal Crack' | 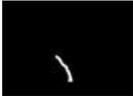 | 1 | 0 | 0.970 | 0.970 |
| 4 | outdoor_drone_00010.jpg | 'Diagonal Crack' | 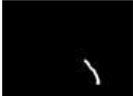 | 1 | 0 | 0.968 | 0.968 |
| 5 | outdoor_drone_00004.jpg | 'Diagonal Crack' | 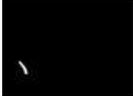 | 1 | 0 | 0.963 | 0.963 |
| 6 | outdoor_drone_00018.jpg | 'Diagonal Crack' | 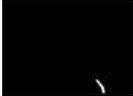 | 1 | 0 | 0.962 | 0.962 |
| 7 | outdoor_drone_00109.jpg | 'Diagonal Crack' | 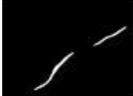 | 2 | 0 | 0.962 | 0.962 |
| 8 | outdoor_drone_00115.jpg | 'Diagonal Crack' | 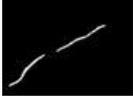 | 2 | 0 | 0.954 | 0.954 |
| 9 | outdoor_drone_00065.jpg | 'Diagonal Crack' | 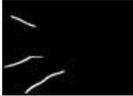 | 3 | 0 | 0.796 | 0.796 |
| 10 | outdoor_drone_00067.jpg | 'Diagonal Crack' | 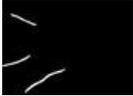 | 3 | 0 | 0.720 | 0.720 |
| 11 | outdoor_drone_00058.jpg | 'Diagonal Crack' | 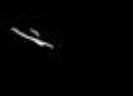 | 3 | 0 | 0.550 | 0.550 |
| 12 | outdoor_drone_00062.jpg | 'Diagonal Crack' | 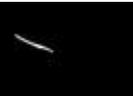 | 1 | 0 | 0.502 | 0.502 |
| 13 | outdoor_drone_00072.jpg | 'Diagonal Crack' | 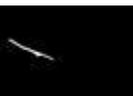 | 2 | 0 | 0.430 | 0.430 |
| 14 | outdoor_drone_00070.jpg | 'Diagonal Crack' | 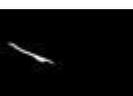 | 2 | 0 | 0.370 | 0.370 |

**Appendix C**

Table 9. Length Estimation of Cracks in Laboratory Drone Images (ground truth 182.5 cm)

| No | Image Name | Estimate Total Length (pixels) | Pixels Per Metric Ratio | Estimate Total Length (cm) | % Error |
|---|---|---|---|---|---|
| 1 | lab_00008.jpg | 611 | 30.55 | 186 | 1.9 |
| 2 | lab_00022.jpg | 168 | 8.40 | 183 | 0.3 |
| 3 | lab_00007.jpg | 169 | 8.45 | 194 | 6.3 |
| 4 | lab_00002.jpg | 727 | 36.35 | 202 | 10.7 |
| 5 | lab_00017.jpg | 669 | 33.45 | 173 | 5.2 |
| 6 | lab_00001.jpg | 595 | 29.75 | 207 | 13.4 |
| 7 | lab_00012.jpg | 609 | 30.45 | 209 | 14.5 |
| 8 | lab_00003.jpg | 185 | 9.25 | 164 | 10.1 |
| 9 | lab_00018.jpg | 604 | 30.20 | 219 | 20.0 |
| 10 | lab_00019.jpg | 173 | 8.65 | 137 | 24.9 |
| 11 | lab_00004.jpg | 659 | 32.95 | 185 | 1.4 |
| 12 | lab_00021.jpg | 136 | 6.80 | 182 | 0.3 |
| 13 | lab_00011.jpg | 150 | 7.50 | 193 | 5.8 |
| 14 | lab_00006.jpg | 516 | 25.80 | 216 | 18.3 |

**References**

[1] M. P. Ismail, "Selection of suitable NDT methods for building inspection," *IOP Conf. Ser. Mater. Sci. Eng.*, vol. 271, p. 012085, Nov. 2017, doi: 10.1088/1757-899X/271/1/012085.

[2] A. M. Paterson, G. R. Dowling, and D. A. Chamberlain, "Building inspection: can computer vision help?," *Autom. Constr.*, vol. 7, no. 1, pp. 13–20, 1997, doi: https://doi.org/10.1016/S0926-5805(97)00031-9.

[3] L. Parente *et al.*, "Image-Based Monitoring of Cracks: Effectiveness Analysis of an Open-Source Machine Learning-Assisted Procedure," *J. Imaging*, vol. 8, no. 2, p. 22, Jan. 2022, doi: 10.3390/jimaging8020022.

[4] B. F. Spencer, V. Hoskere, and Y. Narazaki, "Advances in Computer Vision-Based Civil Infrastructure Inspection and Monitoring," *Engineering*, vol. 5, no. 2, pp. 199–222, Apr. 2019, doi: 10.1016/j.eng.2018.11.030.

[5] *ABECIS*. S.M.A.R.T. Construction Research Group, 2022. Accessed: Apr. 02, 2022. [Online]. Available: https://github.com/SMART-NYUAD/ABECIS

[6] *ABECIS Readme*. S.M.A.R.T. Construction Research Group, 2022. Accessed: Jun. 19, 2022. [Online]. Available: https://github.com/SMART-NYUAD/ABECIS/blob/main/README.md

[7] Y. Z. Ayele, M. Aliyari, D. Griffiths, and E. L. Droguett, "Automatic Crack Segmentation for UAV-Assisted Bridge Inspection," *Energies*, vol. 13, no. 23, p. 6250, Nov. 2020, doi: 10.3390/en13236250.

[8] Z. Y. Wu, R. Kalfarisi, F. Kouyoumdjian, and C. Taelman, "Applying deep convolutional neural network with 3D reality mesh model for water tank crack detection and evaluation," *Urban Water J.*, vol. 17, no. 8, pp. 682–695, Sep. 2020, doi: 10.1080/1573062X.2020.1758166.

[9] J.-H. Lee, S.-S. Yoon, H.-J. Jung, and I.-H. Kim, "Diagnosis of crack damage on structures based on image processing techniques and R-CNN using unmanned aerial vehicle (UAV)," in *Sensors and Smart Structures Technologies for Civil, Mechanical, and Aerospace Systems 2018*, Denver, United States, Mar. 2018, p. 35. doi: 10.1117/12.2296691.

[10] K. He, G. Gkioxari, P. Dollar, and R. Girshick, "Mask R-CNN," 2017, pp. 2961–2969. Accessed: Jun. 19, 2022. [Online]. Available: https://openaccess.thecvf.com/content_iccv_2017/html/He_Mask_R-CNN_ICCV_2017_paper.html

[11] Z. Yu, Y. Shen, and C. Shen, "A real-time detection approach for bridge cracks based on YOLOv4-FPM," *Autom. Constr.*, vol. 122, p. 103514, Feb. 2021, doi: 10.1016/j.autcon.2020.103514.

[12] J. Cen, J. Zhao, X. Xia, and C. Liu, "Application Research on Convolution Neural Network for Bridge Crack Detection," presented at the 2nd International Conference on Computer Engineering, Information Science & Application Technology (ICCIA 2017), Wuhan, China, 2017. doi: 10.2991/iccia-17.2017.24.

[13] J. Jo and Z. Jadidi, "A high precision crack classification system using multi-layered image processing and deep belief learning," *Struct. Infrastruct. Eng.*, vol. 16, no. 2, pp. 297–305, Feb. 2020, doi: 10.1080/15732479.2019.1655068.

[14] Y. Bengio, "Learning Deep Architectures for AI," *Found. Trends® Mach. Learn.*, vol. 2, no. 1, pp. 1–127, 2009, doi: 10.1561/2200000006.

[15] L. A. Silva, H. Sanchez San Blas, D. Peral García, A. Sales Mendes, and G. Villarubia González, "An Architectural Multi-Agent System for a Pavement Monitoring System with Pothole Recognition in UAV Images," *Sensors*, vol. 20, no. 21, p. 6205, Oct. 2020, doi: 10.3390/s20216205.

[16] C. M. Yeum, J. Choi, and S. J. Dyke, "Automated region-of-interest localization and classification for vision-based visual assessment of civil infrastructure," *Struct. Health Monit.*, vol. 18, no. 3, pp. 675–689, May 2019, doi: 10.1177/1475921718765419.

[17] S. Dorafshan, M. Maguire, N. V. Hoffer, and C. Coopmans, "Challenges in bridge inspection using small unmanned aerial systems: Results and lessons learned," in *2017 International Conference on Unmanned Aircraft Systems (ICUAS)*, Miami, FL, USA, Jun. 2017, pp. 1722–1730. doi: 10.1109/ICUAS.2017.7991459.

[18] Y.-J. Cha, W. Choi, and O. Büyüköztürk, "Deep Learning-Based Crack Damage Detection Using Convolutional Neural Networks: Deep learning-based crack damage detection using CNNs," *Comput.-Aided Civ. Infrastruct. Eng.*, vol. 32, no. 5, pp. 361–378, May 2017, doi: 10.1111/mice.12263.

[19] S. Bhowmick, S. Nagarajaiah, and A. Veeraraghavan, "Vision and Deep Learning-Based Algorithms to Detect and Quantify Cracks on Concrete Surfaces from UAV Videos," *Sensors*, vol. 20, no. 21, p. 6299, Nov. 2020, doi: 10.3390/s20216299.

[20] Y. Pan, X. Chen, Q. Sun, and X. Zhang, "Monitoring Asphalt Pavement Aging and Damage Conditions from Low-Altitude UAV Imagery Based on a CNN Approach," *Can. J. Remote Sens.*, vol. 47, no. 3, pp. 432–449, May 2021, doi: 10.1080/07038992.2020.1870217.

[21] A. Bochkovskiy, C.-Y. Wang, and H.-Y. M. Liao, "YOLOv4: Optimal Speed and Accuracy of Object Detection," 2020, doi: 10.48550/ARXIV.2004.10934.

[22] A. Krizhevsky, I. Sutskever, and G. E. Hinton, "ImageNet Classification with Deep Convolutional Neural Networks," *Commun ACM*, vol. 60, no. 6, pp. 84–90, May 2017, doi: 10.1145/3065386.

[23] K. Chen, G. Reichard, X. Xu, and A. Akanmu, "Automated crack segmentation in close-range building façade inspection images using deep learning techniques," *J. Build. Eng.*, vol. 43, p. 102913, Nov. 2021, doi: 10.1016/j.jobe.2021.102913.

[24] H. Zhang, H. Yu, H. Zhang, and W. Yang, "Accurate extraction of cracks on the underside of concrete bridges," in *2017 IEEE International Geoscience and Remote Sensing Symposium (IGARSS)*, 2017, pp. 2341–2344.

[25] D. Dan and Q. Dan, "Automatic recognition of surface cracks in bridges based on 2D-APES and mobile machine vision," *Measurement*, vol. 168, p. 108429, Jan. 2021, doi: 10.1016/j.measurement.2020.108429.

[26] Y. Liu, X. Nie, J. Fan, and X. Liu, "Image-based crack assessment of bridge piers using unmanned aerial vehicles and three-dimensional scene reconstruction," *Comput.-Aided Civ. Infrastruct. Eng.*, vol. 35, no. 5, pp. 511–529, May 2020, doi: 10.1111/mice.12501.

[27] B. Lei, N. Wang, P. Xu, and G. Song, "New Crack Detection Method for Bridge Inspection Using UAV Incorporating Image Processing," *J. Aerosp. Eng.*, vol. 31, no. 5, p. 04018058, Sep. 2018, doi: 10.1061/(ASCE)AS.1943-5525.0000879.

[28] L. Peng and J. Liu, "Detection and analysis of large-scale WT blade surface cracks based on UAV-taken images," *IET Image Process.*, vol. 12, no. 11, pp. 2059–2064, Nov. 2018, doi: 10.1049/iet-ipr.2018.5542.

[29] D. Ribeiro, R. Santos, A. Shibasaki, P. Montenegro, H. Carvalho, and R. Calçada, "Remote inspection of RC structures using unmanned aerial vehicles and heuristic image processing," *Eng. Fail. Anal.*, vol. 117, p. 104813, Nov. 2020, doi: 10.1016/j.engfailanal.2020.104813.

[30] H.-F. Wang, L. Zhai, H. Huang, L.-M. Guan, K.-N. Mu, and G. Wang, "Measurement for cracks at the bottom of bridges based on tethered creeping unmanned aerial vehicle," *Autom. Constr.*, vol. 119, p. 103330, Nov. 2020, doi: 10.1016/j.autcon.2020.103330.

[31] B. J. Perry, Y. Guo, R. Atadero, and J. W. van de Lindt, "Streamlined bridge inspection system utilizing unmanned aerial vehicles (UAVs) and machine learning," *Measurement*, vol. 164, p. 108048, Nov. 2020, doi: 10.1016/j.measurement.2020.108048.

[32] F. Potenza, C. Rinaldi, E. Ottaviano, and V. Gattulli, "A robotics and computer-aided procedure for defect evaluation in bridge inspection," *J. Civ. Struct. Health Monit.*, vol. 10, no. 3, pp. 471–484, Jul. 2020, doi: 10.1007/s13349-020-00395-3.

[33] A. Buatik, "3D MODEL-BASED IMAGE REGISTRATION FOR CHANGE DETECTION IN HISTORICAL STRUCTURES VIA UNMANNED AERIAL VEHICLE," *Int. J. GEOMATE*, vol. 16, no. 58, Jun. 2019, doi: 10.21660/2019.58.8218.

[34] J. Jo, Z. Jadidi, and B. Stantic, "A Drone-Based Building Inspection System Using Software-Agents," in *Intelligent Distributed Computing XI*, vol. 737, M. Ivanović, C. Bădică, J. Dix, Z. Jovanović, M. Malgeri, and M. Savić, Eds. Cham: Springer International Publishing, 2018, pp. 115–121. doi: 10.1007/978-3-319-66379-1_11.

[35] D. Kang and Y.-J. Cha, "Autonomous UAVs for Structural Health Monitoring Using Deep Learning and an Ultrasonic Beacon System with Geo-Tagging: Autonomous UAVs for SHM," *Comput.-Aided Civ. Infrastruct. Eng.*, vol. 33, no. 10, pp. 885–902, Oct. 2018, doi: 10.1111/mice.12375.

[36] Q. Li *et al.*, "Computer vision-based techniques and path planning strategy in a slope monitoring system using unmanned aerial vehicle," *Int. J. Adv. Robot. Syst.*, vol. 17, no. 2, p. 172988142090430, Mar. 2020, doi: 10.1177/1729881420904303.

[37] "Detecting Cracks with AI Technology," *Canon Global*. https://global.canon/en/technology/crack2019.html (accessed Apr. 02, 2022).

[38] fyangneil, *Pavement crack detection: dataset and model.* 2022. Accessed: Apr. 02, 2022. [Online]. Available: https://github.com/fyangneil/pavement-crack-detection

[39] A. Flôr, *arthurflor23/surface-crack-detection.* 2022. Accessed: Apr. 02, 2022. [Online]. Available: https://github.com/arthurflor23/surface-crack-detection

[40] Soni, *Concrete-Crack-Detection.* 2022. Accessed: Apr. 02, 2022. [Online]. Available: https://github.com/vivek6449/Concrete-Crack-Detection

[41] D. Dais, *Crack detection for masonry surfaces.* 2022. Accessed: Apr. 02, 2022. [Online]. Available: https://github.com/dimitrisdais/crack_detection_CNN_masonry

[42] qinnzou, *DeepCrack: Learning Hierarchical Convolutional Features for Crack Detection.* 2022. Accessed: Apr. 02, 2022. [Online]. Available: https://github.com/qinnzou/DeepCrack

[43] "Dynamic – Infrastructure." https://diglobal.tech/ (accessed Jun. 20, 2022).

[44] "T2D2 | Thornton Tomasetti." https://www.thorntontomasetti.com/capability/t2d2 (accessed Jun. 20, 2022).

[45] A. M. Hafiz and G. M. Bhat, "A survey on instance segmentation: state of the art," *Int. J. Multimed. Inf. Retr.*, vol. 9, no. 3, pp. 171–189, Sep. 2020, doi: 10.1007/s13735-020-00195-x.

[46] sliu, *Path Aggregation Network for Instance Segmentation.* 2022. Accessed: Jun. 15, 2022. [Online]. Available: https://github.com/ShuLiu1993/PANet

[47] D. Bolya, *You Only Look At CoefficienTs.* 2022. Accessed: Jun. 15, 2022. [Online]. Available: https://github.com/dbolya/yolact

[48] "facebookresearch/maskrcnn-benchmark: Fast, modular reference implementation of Instance Segmentation and Object Detection algorithms in PyTorch." https://github.com/facebookresearch/maskrcnn-benchmark (accessed Jun. 20, 2022).

[49] "facebookresearch/detectron2: Detectron2 is a platform for object detection, segmentation and other visual recognition tasks." https://github.com/facebookresearch/detectron2 (accessed Apr. 02, 2022).

[50] S. A. Prieto, N. Giakoumidis, and B. García de Soto, "AutoCIS: An Automated Construction Inspection System for Quality Inspection of Buildings," presented at the *38th International Symposium on Automation and Robotics in Construction* (ISARC 2021 Online). November 1-5, 2021, Dubai, UAE. doi: 10.22260/ISARC2021/0090.

[51] "Matrice 300 RTK – Built Tough. Works Smart.," *DJI*. https://www.dji.com/ae/photo (accessed Apr. 02, 2022).

[52] "ACI Concrete Terminology," p. 78.

[53] Ç. F. Özgenel and A. G. Sorguç, "Performance Comparison of Pretrained Convolutional Neural Networks on Crack Detection in Buildings," in *Proceedings of the 35th International Symposium on Automation and Robotics in Construction (ISARC)*, Taipei, Taiwan, Jul. 2018, pp. 693–700. doi: 10.22260/ISARC2018/0094.

[54] L. Pauly, H. Peel, S. Luo, D. Hogg, and R. Fuentes, "Deeper Networks for Pavement Crack Detection," in *Proceedings of the 34th International Symposium on Automation and Robotics in Construction (ISARC)*, Taipei, Taiwan, Jul. 2017, pp. 479–485. doi: 10.22260/ISARC2017/0066.

[55] ayoolaolafenwa, *PixelLib*. 2022. Accessed: Apr. 02, 2022. [Online]. Available: https://github.com/ayoolaolafenwa/PixelLib

[56] *Faster R-CNN and Mask R-CNN in PyTorch 1.0*. Meta Research, 2022. Accessed: Jun. 14, 2022. [Online]. Available: https://github.com/facebookresearch/maskrcnn-benchmark

[57] K. Wada, *Labelme: Image Polygonal Annotation with Python*. 2022. doi: 10.5281/zenodo.5711226.

[58] *LabelMe annotation tool source code*. MIT CSAIL Computer Vision, 2022. Accessed: Jun. 20, 2022. [Online]. Available: https://github.com/CSAILVision/LabelMeAnnotationTool

[59] P. Ko, B. García de Soto, and S.A. Prieto, "ABECIS-Crack-Segmentation-Dataset." 2022. doi: 10.7910/DVN/QYGDMM.

[60] P. Ko, S. A. Prieto, and B. García de Soto, "ABECIS: an Automated Building Exterior Crack Inspection System using UAVs, Open-Source Deep Learning and Photogrammetry," presented at the *38th International Symposium on Automation and Robotics in Construction* (ISARC 2021 Online). November 1-5, 2021, Dubai, UAE. doi: 10.22260/ISARC2021/0086.

[61] *detectron2*. Meta Research, 2022. Accessed: Apr. 01, 2022. [Online]. Available: https://github.com/facebookresearch/detectron2

[62] "Loss Functions — ML Glossary documentation." https://ml-cheatsheet.readthedocs.io/en/latest/loss_functions.html (accessed Apr. 02, 2022).

[63] "Classification: Accuracy | Machine Learning Crash Course," *Google Developers*. https://developers.google.com/machine-learning/crash-course/classification/accuracy (accessed Apr. 02, 2022).

[64] "PyQt6 · PyPI." https://pypi.org/project/PyQt6/ (accessed Apr. 02, 2022).

[65] "opencv-python · PyPI." https://pypi.org/project/opencv-python/ (accessed Apr. 02, 2022).

[66] "ABECIS/README.md at main · SMART-NYUAD/ABECIS." https://github.com/SMART-NYUAD/ABECIS/blob/main/README.md (accessed Jun. 20, 2022).

[67] M. R. Jahanshahi, S. F. Masri, C. W. Padgett, and G. S. Sukhatme, "An innovative methodology for detection and quantification of cracks through incorporation of depth perception," *Mach. Vis. Appl.*, vol. 24, no. 2, pp. 227–241, Feb. 2013, doi: 10.1007/s00138-011-0394-0.

[68] F. Y. Shih, *Image processing and mathematical morphology: fundamentals and applications*. 2017.

[69] P. Ko, *ABECIS-Results*. 2022. Accessed: Jun. 14, 2022. [Online]. Available: https://github.com/SMART-NYUAD/ABECIS-Results

[70] "OpenCV: Canny Edge Detection." https://docs.opencv.org/4.x/da/d22/tutorial_py_canny.html (accessed Apr. 02, 2022).

[71] H. Yu, W. Yang, H. Zhang, and W. He, "A UAV-based crack inspection system for concrete bridge monitoring," in *2017 IEEE International Geoscience and Remote Sensing Symposium (IGARSS)*, 2017, pp. 3305–3308.

[72] J. Rau, K. Hsiao, J. Jhan, S. Wang, W. Fang, and J. Wang, "Bridge crack detection using multi-rotary UAV and object-base image analysis," *Int. Arch. Photogramm. Remote Sens. Spat. Inf. Sci.*, vol. 42, p. 311, 2017.

[73] S. Dorafshan, R. J. Thomas, and M. Maguire, "Comparison of deep convolutional neural networks and edge detectors for image-based crack detection in concrete," *Constr. Build. Mater.*, vol. 186, pp. 1031–1045, Oct. 2018, doi: 10.1016/j.conbuildmat.2018.08.011.

[74] D. Han, "Crack detection of UAV concrete surface images," in *Applications of Machine Learning*, San Diego, United States, Sep. 2019, p. 37. doi: 10.1117/12.2525174.

[75] V. Barrile, G. Bilotta, and A. Nunnari, "UAV and Computer Vision, Detection of Infrastructure Losses and 3D Modeling," *ISPRS Ann. Photogramm. Remote Sens. Spat. Inf. Sci.*, vol. 4, p. 135, 2017.

[76] D. Kang and Y.-J. Cha, "Damage detection with an autonomous UAV using deep learning," in *Sensors and Smart Structures Technologies for Civil, Mechanical, and Aerospace Systems 2018*, 2018, vol. 10598, pp. 7–14.

[77] S. Dorafshan, R. J. Thomas, and M. Maguire, "Fatigue Crack Detection Using Unmanned Aerial Systems in Fracture Critical Inspection of Steel Bridges," *J. Bridge Eng.*, vol. 23, no. 10, p. 04018078, Oct. 2018, doi: 10.1061/(ASCE)BE.1943-5592.0001291.

[78] A. B. Ersoz, O. Pekcan, and T. Teke, "Crack identification for rigid pavements using unmanned aerial vehicles," in *IOP Conference Series: Materials Science and Engineering*, 2017, vol. 236, no. 1, p. 012101.

[79] B. Ramalingam *et al.*, "Visual Inspection of the Aircraft Surface Using a Teleoperated Reconfigurable Climbing Robot and Enhanced Deep Learning Technique," *Int. J. Aerosp. Eng.*, vol. 2019, pp. 1–14, Sep. 2019, doi: 10.1155/2019/5137139.

[80] Y. Liu, J. K. W. Yeoh, and D. K. H. Chua, "Deep Learning–Based Enhancement of Motion Blurred UAV Concrete Crack Images," *J. Comput. Civ. Eng.*, vol. 34, no. 5, p. 04020028, Sep. 2020, doi: 10.1061/(ASCE)CP.1943-5487.0000907.